\documentclass[preprint,12pt]{elsarticle}

\pdfoutput=1
\usepackage{amssymb}
\usepackage{bm}
\usepackage{algorithm}
\usepackage{algorithmic}
\usepackage{amsthm}

\usepackage{amssymb}
\usepackage{amsmath}
\usepackage{graphicx}
\usepackage{subfigure}
\usepackage{booktabs}

\usepackage[]{caption2}

\journal{arxiv}

\begin{document}

\begin{frontmatter}



\title{Sampling-based Causal Inference in Cue Combination and its Neural Implementation}


\author[thu]{Zhaofei Yu}
\author[thu,usc]{Feng Chen}
\author[thu]{Jianwu Dong}
\author[thu]{Qionghai Dai}

\address[thu]{Center for Brain-Inspired Computing Research,\\
Department of Automation, Tsinghua University, Beijing 100084, China}
\address[usc]{Beijing Key Laboratory of Security in Big Data Processing and Application,\\
 Beijing 100084, China}

\begin{abstract}
Causal inference in cue combination is to decide whether the cues have a single cause or multiple causes. Although the Bayesian causal inference model explains the problem of causal inference in cue combination successfully, how causal inference in cue combination could be implemented by neural circuits, is unclear. The existing method based on calculating log posterior ratio with variable elimination has the problem of being unrealistic and task-specific. In this paper, we take advantages of the special structure of the Bayesian causal inference model and propose a hierarchical inference algorithm based on importance sampling. A simple neural circuit is designed to implement the proposed inference algorithm. Theoretical analyses and experimental results demonstrate that our algorithm converges to the accurate value as the sample size goes to infinite. Moreover, the neural circuit we design can be easily generalized to implement inference for other problems, such as the multi-stimuli cause inference and the same-different judgment.
\end{abstract}

\begin{keyword}
Causal inference \sep importance sampling\sep cue combination\sep  neural circuit


\end{keyword}

\end{frontmatter}


\section{Introduction}
Human brain receives cues from multiple sensory modalities and integrates them in an optimal way \cite{pouget2013probabilistic}. The cues from the outside world are noisy observations of
stimuli reflecting uncertainty. It has been demonstrated that, if all cues have the same cause, the optimal process of cue combination is a process
 of Bayesian inference \cite{alais2004ventriloquist,ernst2002humans,knill2003humans,van1996humans}. However, the truth is that, we receive information from various sources simultaneously in our daily life, which means the cues
  may come from different causes. How to decide whether a single cause or multiple causes is responsible for the cues, known as causal inference
   in cue combination, is an important problem. This problem is the precondition of cue combination and is quite common in our daily life \cite{seilheimer2014models,ursino2014neurocomputational}. For example, at a cocktail party, we need to decide whether the face and voice belong to the person who calls our name \cite{kayser2015multisensory}.
 Recently, the problem of causal inference in cue combination is partially answered by Kording et al.\cite{kording2007causal} and Sato et al.\cite{sato2007bayesian}, who propose the Bayesian
   causal inference model. Their causal inference model successfully explains the problem of causal inference in cue combination. Yet, how causal
   inference in cue combination could be implemented by neural circuit, is unclear. Solving this problem benefits not only theoretical researches
   but also practical applications. On the one hand, causal inference is the basis for cue combination. On the other hand, if the causal
 inference could be implemented by neural circuits, the neural circuits could be used to perform causal inference in cue combination for robots.

Over the past decade, several methods with different probability codes have been proposed to perform probability inference with neural circuits. Rao \cite{rao2004bayesian,rao2004hierarchical,doya2007bayesian} establishes the relationship between the dynamic equation of neural circuits and the inference of probabilistic graphical models. He proves that the process that the firing rate of neurons in the recurrent neural circuit varies with respect to time is a process of posterior probabilities inference in a hidden Markov model, under the condition that the firing rate is proportional to the log of posterior probabilities. Ott and Stoop \cite{ott2006neurodynamics} build the relationship between the dynamical equation of continuous Hopfield network and belief propagation on a binary Markov random field. Sampling is another commonly accepted way to perform inference by neural circuits. Based on Monte Carlo sampling, Huang and Rao \cite{huang2014neurons} build a spiking network model to perform approximate inference for any hidden Markov model. Maass et al.\cite{buesing2011neural,pecevski2011probabilistic,legenstein2014ensembles}
propose that stochastic networks of spiking neurons could implement inference for graphical models by Markov chain Monte Carlo. Shi and Griffiths \cite{shi2009neural} apply importance sampling to perform inference of chain Bayesian model and design neural circuits to implement it. Another important framework is Probabilistic population coding (PPC), the core idea of which is that the neurons are encoders of distributions, instead of the values of variables \cite{ma2006bayesian,ma2014neural,orhan2015neural}. Ma et al.\cite{ma2006bayesian} present that the inference of cue integration can be conducted simply by linear combinations of each population activity with PPC. The method is exploited thereafter by Beck et al.\cite{beck2008probabilistic} to realize the Bayesian decision making and the inference of marginalization \cite{beck2011marginalization}.

To the best of our knowledge, the only work implementing causal inference in cue combination with neural circuits is proposed by Ma et al. \cite{ma2013towards} in 2013. They calculate the ratio of the posterior probabilities of both situations (a single cause or multiple causes) with variable elimination and then design a neural circuit to implement it. This method suffers from three shortcomings. Firstly, the circuits they design are task-specific and only work on two stimuli. If we want to implement multi-stimuli causal inference \cite{wozny2008human} with the same method, the circuit will be completely different. What's more, the required number of operations increases faster than linear with respect to the number of stimuli, which makes the neural circuit unrealistic \cite{ma2013towards}. Secondly, it is hard to generalize the circuit to implement a similar task called same-different judgment \cite{van2012optimal}. Thirdly, since how to implement logarithmic operations with neurons remains unknown, approximations are taken in their neural circuit so that they could only get near-optimal results.

In this paper, different from calculating the posterior ratio with variable elimination in \cite{ma2013towards}, we propose a hierarchical inference algorithm based on importance sampling, which takes advantages of the special structure of the causal inference model.  A neural circuit with hierarchical structure is then designed corresponding to the bottom-up inference process. The proposed method has three advantages. Firstly, the neural circuit is simple and it is easy to be realized by PPC and some simple plausible neural operations. Secondly, it is easy to generalize this neural circuit to implement inference for other problems, such as the multi-stimuli cause inference and the same-different judgment. Thirdly, a theoretical proof is given that the sampling-based method converges to the accurate value with probability one as sample size tends to infinity.

The rest of this paper is organized as follows. Section 2 briefly reviews the causal inference in cue combination. In section 3 we present a sampling-based inference algorithm and design the corresponding neural circuit. The experimental results are shown in section 4. We generalize our method to solve other two problems in section 5 and make a conclusion in section 6.

\section{The Causal Inference Model In Cue Combination}
\begin{figure}
\centering
\includegraphics{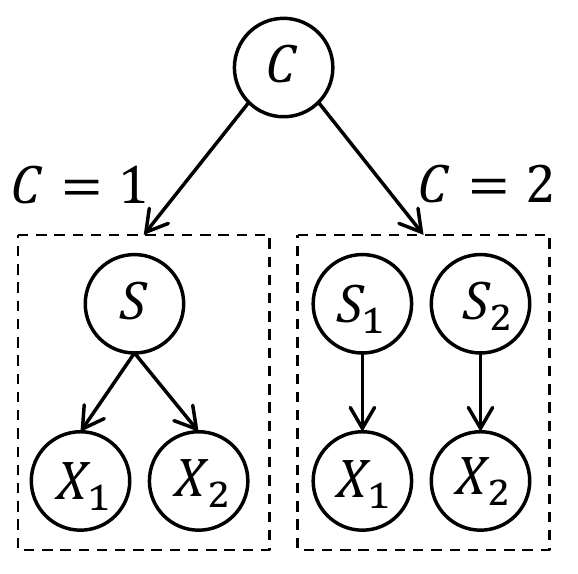}
\caption{The causal inference model in cue combination.}
\label{fig_1}
\end{figure}

The problem of causal inference in cue combination is to infer whether cues come from a single or multiple causes.
Kording et al. \cite{kording2007causal} and Sato et al. \cite{sato2007bayesian} propose a causal inference model of cue combination respectively, which could explain physiological and psychological experiments successfully. Here, we briefly review this model and the stimuli considered here only include visual and auditory ones. The multi-stimuli problem will be explained in section 5. In Fig. \ref{fig_1}, node $C$ represents the common-cause variable, $S$, ${S_1}$, and ${S_2}$ express the stimuli. ${X_1}$ and ${X_2}$ are cues received by the sensory system. The state of cause $C$ is 1 or 2, where $C=1$ means the cues have the same cause and  $C=2$ means the cues have two different causes. For simplicity, we assume that $P\left( {C = 1} \right)$ is equal to $P\left( {C = 2} \right)$, both of which have a probability 0.5. When $C=1$ , there is a stimulus $S$ with distribution $P(S)$ corresponding to the common cause, where $P(S)$ is a Gaussian distribution with mean 0 and variance $\sigma _S^2$. Two measurements ${X_1}$ and ${X_2}$ are generated from two Gaussian distributions with different variances $\sigma _1^2$ and $\sigma _2^2$, but with the same mean $S$. When $C=2$, there are two different stimuli ${S_1}$ and ${S_2}$, which are drawn from the same Gaussian distribution with mean 0 and variance $\sigma _S^2$. Then two measurements ${X_1}$ and ${X_2}$ are drawn from two different Gaussian distributions with their means being ${S_1}$ and ${S_2}$, and their variances being $\sigma _1^2$ and $\sigma _2^2$ respectively. Based on the definitions above, the causal inference problem is to decide whether $C = 1$  or $C = 2$  according to the measurements ${X_1}$ and ${X_2}$.

\section{Sampling-Based Causal Inference}
In this section, we first convert the causal inference model to a three-layer Bayesian network. Then we propose a sampling-based hierarchical inference method and design the corresponding neural circuit. We demonstrate that this circuit can be realized by PPC and simple plausible neural operations.

\subsection{The three-layer Bayesian network model}
\begin{figure}
\centering
\includegraphics{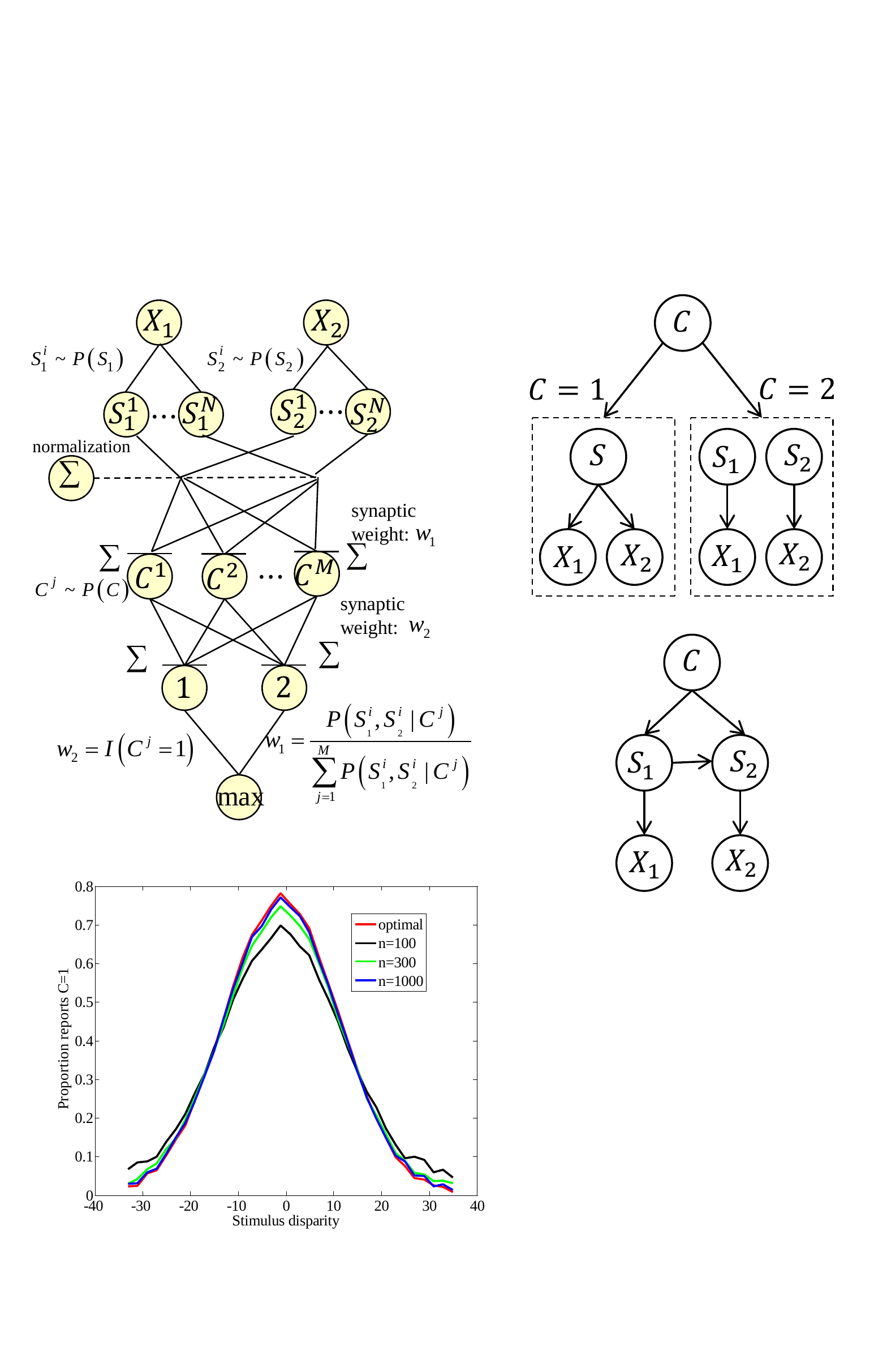}
\caption{ The three-layer Bayesian network equivalent to the causal inference model in Fig. 1.}
\label{fig_2}
\end{figure}
In this paper, the problem is to infer the state of node C. In order to simplify inference, we convert the causal inference model above to a three-layer Bayesian model (Fig. \ref{fig_2}) with some appropriate prior probabilities and conditional probabilities. In the new model, node C is the common-cause variable, which is similar to that in the causal inference model. ${S_1}$ and ${S_2}$ refer to two different stimuli, such as visual and auditory stimuli. The conditional probability of ${S_1}$ and ${S_2}$ under $C$ is expressed as $P\left( {{S_1},{S_2}|C} \right)$. We define $P\left( {{S_1},{S_2}|C = 1} \right) = \delta \left( {{S_1} - {S_2}} \right)\frac{1}{{\sqrt {2\pi } {\sigma _S}}}\exp \left( { - \frac{{S_1^2}}{{2\sigma _S^2}}} \right)$ and $P\left( {{S_1},{S_2}|C = 2} \right) = \frac{1}{{2\pi \sigma _S^2}}\exp \left( { - \frac{{S_1^2 + S_2^2}}{{2\sigma _S^2}}} \right)$, where $\delta \left( {{S_1} - {S_2}} \right)$ is the Dirac Delta distribution. ${X_1}$ and ${X_2}$ are measurements from ${S_1}$ and ${S_2}$ respectively. The conditional probability of ${X_1}$ under ${S_1}$ is defined by $P\left( {{X_1}|{S_1}} \right) = \frac{1}{{\sqrt {2\pi } {\sigma _1}}}\exp \left( { - \frac{{{{\left( {{X_1} - {S_1}} \right)}^2}}}{{2\sigma _1^2}}} \right)$ and the conditional probability of ${X_2}$ under ${S_2}$ is defined by $P\left( {{X_2}|{S_2}} \right) = \frac{1}{{\sqrt {2\pi } {\sigma _2}}}\exp \left( { - \frac{{{{\left( {{X_2} - {S_2}} \right)}^2}}}{{2\sigma _2^2}}} \right)$. It is easy to verify that this Bayesian network is equivalent to the causal inference model in
 Fig. \ref{fig_1}.

\subsection{Sampling-based inference algorithm}
Several methods have been developed to perform inference for the Bayesian model shown in Fig. \ref{fig_2}, such as belief propagation (BP) \cite{koller2009probabilistic} and Markov chain Monte Carlo (MCMC) \cite{andrieu2003introduction}, all of which are able to be implemented by neural circuits \cite{steimer2009belief,litvak2009cortical,buesing2011neural,pecevski2011probabilistic}. However, all the circuits have the shortcoming of being task-specific. Specifically, the neural circuit for belief propagation \cite{steimer2009belief,litvak2009cortical} requires pools of spiking neurons to represent function nodes of the factor graph. It is hard to generalize the circuit of the Bayesian model in Fig. \ref{fig_2} to implement multi-stimuli causal inference. Similarly, the neural circuit \cite{buesing2011neural,pecevski2011probabilistic} based on MCMC should meet the neural computability condition (NCC) and the circuit will be completely different for multi-stimuli causal inference. In this paper, we aim to build a general-purpose neural circuit for causal inference in cue combination. Here we utilize importance sampling to perform inference. Importance sampling is a kind of Monte Carlo methods in statistics, which is used to estimate the intractable integrals by random sampling. Different from other Monte Carlo methods, importance sampling generates samples from a simple distribution rather than the original distribution \cite{hachiya2012importance,cheng2000ais}. Here we give a simple example.
\begin{equation}
\label{eq_1}
\begin{array}{l}
\;\;\;E{\left( {f\left( X \right)} \right)_{P\left( X \right)}}\\
 = \int\limits_X {f\left( X \right)P\left( X \right)} dX\\
 = \int\limits_X {\frac{{f\left( X \right)P\left( X \right)}}{{g\left( X \right)}}g\left( X \right)dX} \\
 = E{\left( {\frac{{f\left( X \right)P\left( X \right)}}{{g\left( X \right)}}} \right)_{g\left( X \right)}}\\
\approx \frac{1}{m}\sum\limits_{\scriptstyle i = 1:\atop
\scriptstyle{X_i} \sim g\left( X \right)}^m {\frac{{f\left( {{X_i}} \right)P\left( {{X_i}} \right)}}{{g\left( {{X_i}} \right)}}}
\end{array}
\end{equation}

In equation (\ref{eq_1}), the goal is to calculate the mathematical expectation of $f\left( X \right)$, where $X$ follows the distribution $P\left( X \right)$. There are cases where we can't sample from the original distribution $P\left( X \right)$ of variable $X$ directly. Instead, we can calculate the expectation of $f\left( X \right)P\left( X \right)/g\left( X \right)$ with $X$ following the simple distribution $g\left( X \right)$.
Note that for the region with larger value of  $g\left( X \right)$, the sampling points should be denser, which means the samples are more important.

Using importance sampling to perform inference has its neural basis. The responses of neurons have been interpreted as Monte Carlo samples by Hoyer and Hyvarinen \cite{hoyer2003interpreting}, which means that the state of each neuron is drawn randomly from a special distribution. Shi and Griffiths \cite{shi2009neural} have used importance sampling to perform inference of chain Bayesian network. By taking in the idea of hierarchical inference, we generalize importance sampling to our model. What\rq s more, we will prove the convergence of the sampling-based inference method. We first consider a Bayesian network with only two nodes $A$ and $B$, where $A$ is the parent node of $B$. It is easy to obtain the conditional expectation of $A$ given $B$ with importance sampling:

\begin{equation}
\label{eq_2}
\begin{array}{l}
E\left( {f\left( A \right)|B} \right) = \sum\limits_A {f\left( A \right)P\left( {A|B} \right)}  = \frac{{\sum\limits_A {f\left( A \right)P\left( {A|B} \right)P\left( B \right)} }}{{P\left( B \right)}}\;\\
\;\;\;\;\;\;\;\;\;\;\;\;\;\;\;\;\;\; = \frac{{\sum\limits_A {f\left( A \right)P\left( {B|A} \right)P\left( A \right)} }}{{\sum\limits_A {P\left( {B|A} \right)P\left( A \right)} }} = \frac{{E{{\left( {f\left( A \right)P\left( {B|A} \right)} \right)}_{P\left( A \right)}}}}{{E{{\left( {P\left( {B|A} \right)} \right)}_{P\left( A \right)}}}}\\
\;\;\;\;\;\;\;\;\;\;\;\;\;\;\;\;\;\; \approx \frac{{\sum\limits_{{A^i}:{A^i} \sim P\left( A \right)} {f\left( {{A^i}} \right)P\left( {B|{A^i}} \right)} }}{{\sum\limits_{{A^i}:{A^i} \sim P\left( A \right)} {P\left( {B|{A^i}} \right)} }}\\
\;\;\;\;\;\;\;\;\;\;\;\;\;\;\;\;\;\; = \sum\limits_{{A^i}:{A^i} \sim P\left( A \right)} {f\left( {{A^i}} \right)\frac{{P\left( {B|{A^i}} \right)}}{{\sum\limits_{{A^i}:{A^i} \sim P\left( A \right)} {P\left( {B|{A^i}} \right)} }}}
\end{array}
\end{equation}

In equation (\ref{eq_2}), the approximation holds as we use importance sampling to estimate the expectation. ${A^i} \sim P\left( A \right)$ means that the sample ${A^i}$ is drawn from the distribution $P\left( A \right)$. Here $A$ is discrete and the sums are replaced by integrals when $A$ is continuous. It should be noted that the same samples of ${A^i}$ are used in both sum. When we calculate the last term in (\ref{eq_2}), we first calculate the sum of ${A^i}$ in the denominator and then calculate the sum of ${A^i}$ in the numerator. Equation (\ref{eq_2}) could be generalized to solve the inference problem in Fig. \ref{fig_2}:

\begin{equation}
\label{eq_3}
\begin{array}{l}
\;\;\;P\left( {C = 1|{X_1} = {x_1},{X_2} = {x_2}} \right)\\
 = \;P\left( {C = 1|{x_1},{x_2}} \right)\\
 = \int\limits_{{S_1},{S_2}} {P\left( {C = 1,{S_1},{S_2}|{x_1},{x_2}} \right)d{S_1},{S_2}} \\
 = \int\limits_{{S_1},{S_2}} {P\left( {C = 1|{S_1},{S_2}} \right)P\left( {{S_1},{S_2}|{x_1},{x_2}} \right)d{S_1},{S_2}} \\
 = \frac{{\int\limits_{{S_1},{S_2}} {P\left( {C = 1|{S_1},{S_2}} \right)P\left( {{x_1},{x_2}|{S_1},{S_2}} \right)P\left( {{S_1},{S_2}} \right)d{S_1},{S_2}} }}{{\int\limits_{{S_1},{S_2}} {P\left( {{x_1},{x_2}|{S_1},{S_2}} \right)P\left( {{S_1},{S_2}} \right)d{S_1},{S_2}} }}\\
 = \frac{{E{{\left( {P\left( {C = 1|{S_1},{S_2}} \right)P\left( {{x_1},{x_2}|{S_1},{S_2}} \right)} \right)}_{P\left( {{S_1},{S_2}} \right)}}}}{{E{{\left( {P\left( {{x_1},{x_2}|{S_1},{S_2}} \right)} \right)}_{P\left( {{S_1},{S_2}} \right)}}}}\\
 \approx \sum\limits_{\scriptstyle i = 1\atop
\scriptstyle S_{_1}^i,S_{_2}^i \sim P\left( {{S_1},{S_2}} \right)}^N {P\left( {C = 1|S_{_1}^i,S_{_2}^i} \right)\frac{{P\left( {{x_1},{x_2}|S_{_1}^i,S_{_2}^i} \right)}}{{\sum\limits_{\scriptstyle i = 1\atop
\scriptstyle S_{_1}^i,S_{_2}^i \sim P\left( {{S_1},{S_2}} \right)}^N {P\left( {{x_1},{x_2}|S_{_1}^i,S_{_2}^i} \right)} }}} \\
 = \sum\limits_{\scriptstyle i = 1\atop
\scriptstyle S_{_1}^i,S_{_2}^i \sim P\left( {{S_1},{S_2}} \right)}^N {\frac{{P\left( {S_{_1}^i,S_{_2}^i|C = 1} \right)}}{{P\left( {S_{_1}^i,S_{_2}^i|C = 1} \right) + P\left( {S_{_1}^i,S_{_2}^i|C = 2} \right)}}\frac{{P\left( {{x_1},{x_2}|S_{_1}^i,S_{_2}^i} \right)}}{{\sum\limits_{\scriptstyle i = 1\atop
\scriptstyle S_{_1}^i,S_{_2}^i \sim P\left( {{S_1},{S_2}} \right)}^N {P\left( {{x_1},{x_2}|S_{_1}^i,S_{_2}^i} \right)} }}} \\
 = \sum\limits_{\scriptstyle i = 1\atop
\scriptstyle S_{_1}^i,S_{_2}^i \sim P\left( {{S_1},{S_2}} \right)}^N {I\left( {S_{_1}^i = S_{_2}^i} \right)\frac{{P\left( {{x_1},{x_2}|S_{_1}^i,S_{_2}^i} \right)}}{{\sum\limits_{\scriptstyle i = 1\atop
\scriptstyle S_{_1}^i,S_{_2}^i \sim P\left( {{S_1},{S_2}} \right)}^N {P\left( {{x_1},{x_2}|S_{_1}^i,S_{_2}^i} \right)} }}}
\end{array}
\end{equation}

In equation (\ref{eq_3}), the sample $S_{_1}^i,S_{_2}^i$ is drawn from $P\left( {{S_1},{S_2}} \right)$. We abbreviate ${X_1} = {x_1},{X_2} = {x_2}$ to ${x_1},{x_2}$ and this will hold in the rest of the paper. ${I\left( {S_{_1}^i = S_{_2}^i} \right)}$ is an indicator function, it equals to 1 only when $S_1^i = S_2^i$. The last equality holds due to the definitions of $P\left( {{S_1},{S_2}|C = 1} \right)$ and $P\left( {{S_1},{S_2}|C = 2} \right)$. Note that equation  (\ref{eq_3}) also holds for $P\left( {C = 1} \right) \ne 0.5$. It is easy to find that equation (\ref{eq_3}) remains the hierarchical structure of Bayesian model in Fig. \ref{fig_2}. Based on this, a neural circuit with a hierarchical structure could be designed corresponding to the bottom-up process of inference. We will discuss this part in detail in the following subsection.

An important index for sampling-based algorithm is its accuracy. The following theorem elucidates that our algorithm converges to the accurate value with probability one as the sample size goes to infinity. The proof of theorem 1 is provided in Appendix A.

\textbf{Theorem 1.} The distributions $P\left( C \right)$, $P\left( {{S_1},{S_2}|C} \right)$, $P\left( {{X_1}|{S_1}} \right)$ and
$P\left( {{X_2}|{S_1}} \right)$ are defined on the Bayesian network in Fig.\ref{fig_2}. $S_{_1}^i,S_{_2}^i\sim P\left( {{S_1},{S_2}} \right)$, then for arbitrary small number ${{\varepsilon}}$\\
\begin{equation}
\label{eq_4}
\begin{array}{l}
\mathop {\lim }\limits_{N \to \infty } P\left( {\left| {\sum\limits_{i = 1}^N {P\left( {C = 1|S_{_1}^i,S_{_2}^i} \right)\frac{{P\left( {{x_1},{x_2}|S_{_1}^i,S_{_2}^i} \right)}}{{\sum\limits_{i = 1}^N {P\left( {{x_1},{x_2}|S_{_1}^i,S_{_2}^i} \right)} }} - P\left( {C = 1|{x_1},{x_2}} \right)} } \right| < \varepsilon } \right) = 1
\end{array}
\end{equation}

\subsection{Implementation with Neural Circuits}
\begin{figure}
\centering
\includegraphics{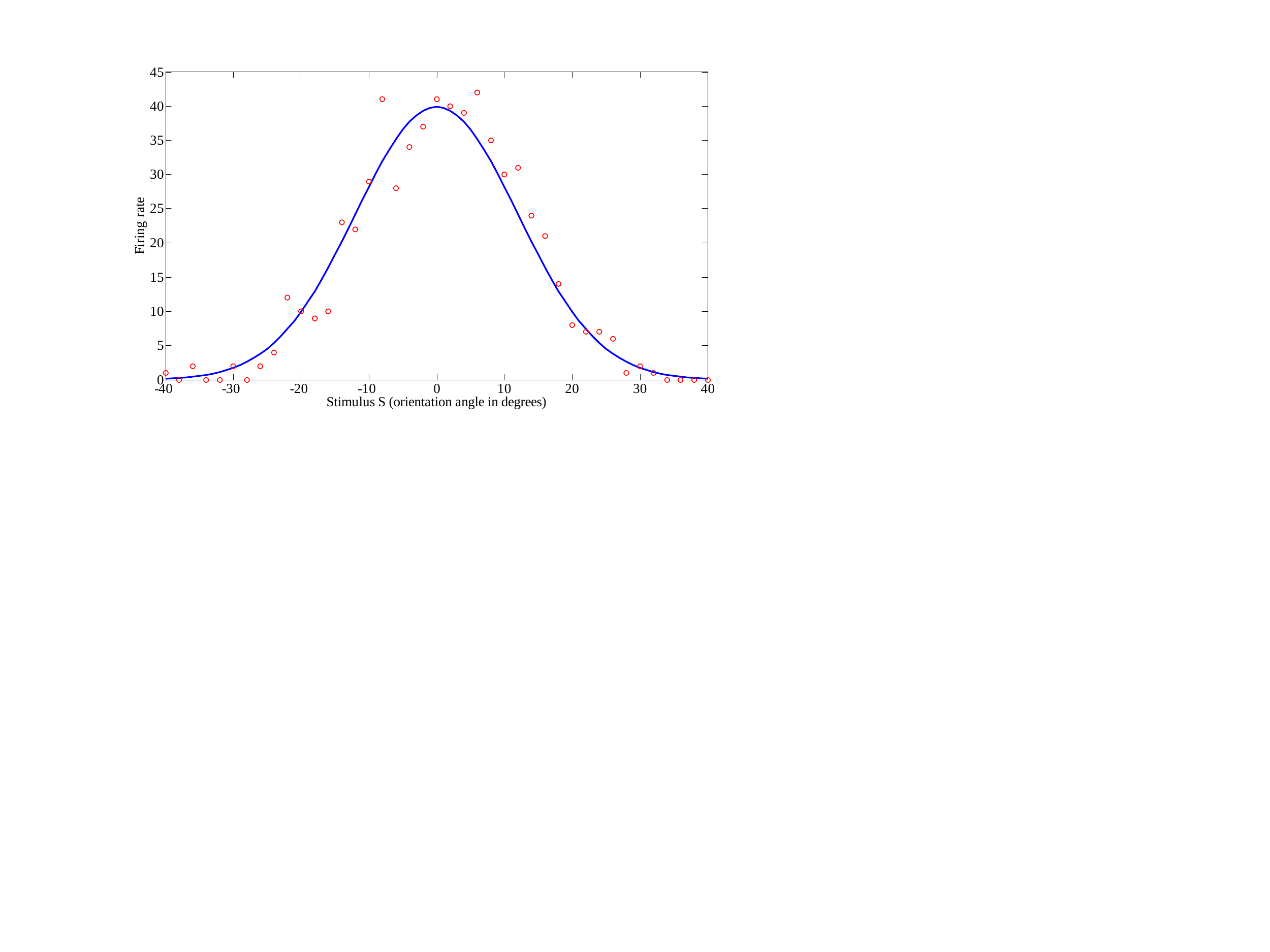}
\caption{The tuning curve of a neuron  in the primary visual cortex (V1) and its neural variability.}
\label{fig_3}
\end{figure}

\begin{figure}
\centering
\includegraphics{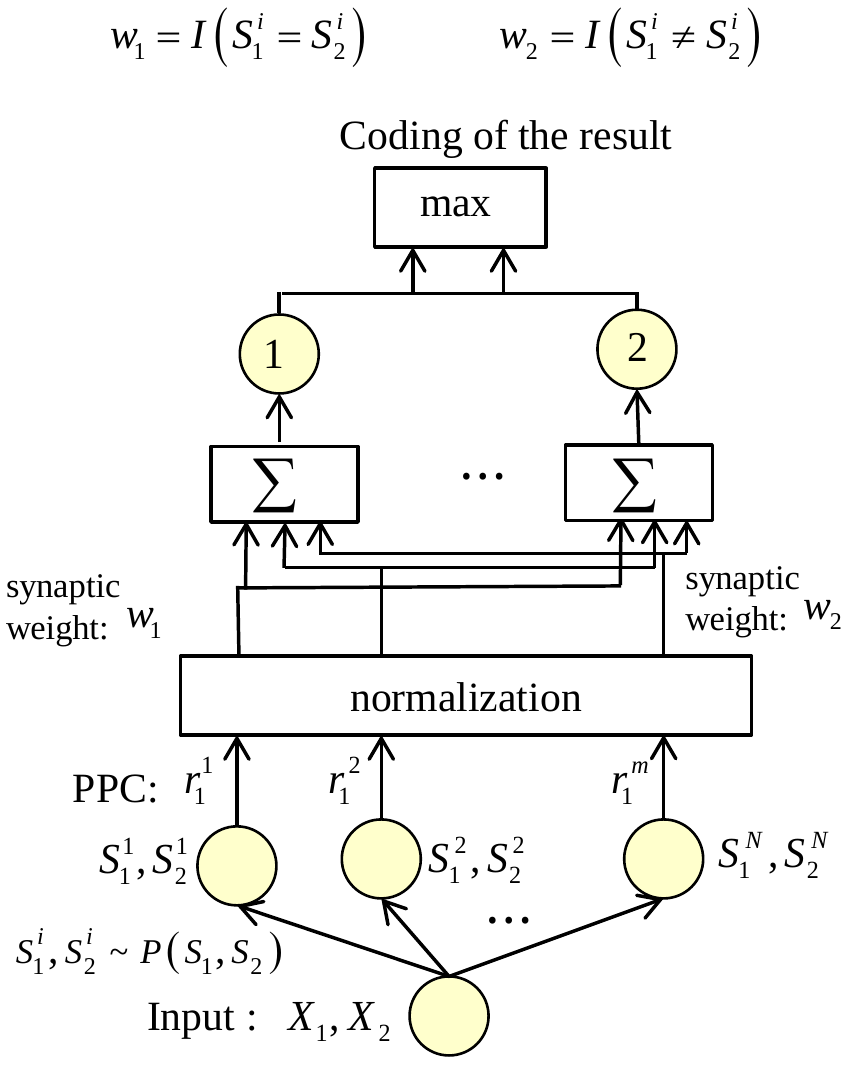}
\caption{The neural circuit of sampling-based method for causal inference in cue combination.}
\label{fig_4}
\end{figure}

In this section, we design a neural circuit to implement sampling-based causal inference in cue combination. According to recent studies, one of the most accepted neural circuits to implement probability inference is based on PPC and some plausible neural operations. Related researches include that of Ma et al.\cite{ma2006bayesian}, presenting that the inference of cue integration can be conducted by PPC and a plausible neural operation-linear combinations. Beck et al.\cite{beck2011marginalization} implement the inference of marginalization with PPC, quadratic nonlinearity and divisive normalization. Our method adopts the same structure as that of Ma and Beck, and the neural circuit is designed based on PPC and three types of plausible neural operations, including multiplication, normalization and linear combinations.
Here we first give a simple explanation of PPC. PPC takes advantages of the variability in neuronal responses and considers a population of neurons as the encoders of probability distributions, rather than the values of variables. Specifically, for independent Poisson spiking neurons, the distribution of the responses $r = \left\{ {{r_1},{r_2},...,{r_N}} \right\}$ to the input stimulus $S$ is $P\left( {r|S} \right) = \prod\limits_i {\frac{{{e^{ - {f_i}\left( s \right)}}{f_i}{{\left( s \right)}^{{r_i}}}}}{{{r_i}!}}} $, where ${f_i}\left( s \right)$
is the tuning curve of the neuron $i$.  The tuning curve is a function of $S$, which represents the average firing rate to stimulus $S$ over trials (In theory, an infinite number of trials). Fig. \ref{fig_3} shows an example, the blue curve is the Gaussian-like tuning curve of a neuron in the primary visual cortex (V1). This neuron is sensitive to the moving direction of the stimulus. The red circles are the firing rates with respect to different moving direction of the stimulus in a trial. The red circles are not always on the blue curve because the tuning curve is the average firing rate and neural response has variability. The neural circuits for equation (\ref{eq_3}) are shown in Fig. \ref{fig_4}. We suppose that there are Poisson spiking neurons $S_1^1S_2^1$, $S_1^2S_2^2$,..., $S_1^NS_2^N$ with their states sampling from $P\left( {{S_1},{S_2}} \right)$. The tuning curve of the neuron $S_1^iS_2^i$ is proportional to $P\left( {{X_1},{X_2}|S_1^i,S_2^i} \right)$. These assumptions are reasonable as physiologically studies \cite{de1982orientation,coppola1998unequal,furmanski2000oblique} have demonstrated that the quantity of neurons in human brain follows some prior distributions. The Poisson spiking neurons $S_1^1S_2^1$, $S_1^2S_2^2$,..., $S_1^NS_2^N$ are used to code the input stimuli ${X_1}$, ${X_2}$ and the output firing rates are $r_1^1$, $r_1^2$,..., $r_1^N$ respectively. The firing rates are then normalized. Note that the normalization operation can be realized by inhibitory neurons \cite{kouh2008canonical}. If we use $R$ to express the total firing rate, where $R = \sum\limits_i {{r_i}} $, then we can get $E\left( {{r_i}/R|R = n} \right) = P\left( {{X_1},{X_2}|S_1^i,S_2^i} \right)/ \left( {\sum\limits_i {P\left( {{X_1},{X_2}|S_1^i,S_2^i} \right)} } \right)$, which is proved in \cite{shi2009neural}. The equation above means the expectation of the normalized firing rate for Poisson neurons equals to normalized probability $P\left( {{X_1},{X_2}|S_{_1}^i,S_{_2}^i} \right)/ \left( {\sum\limits_{i = 1}^N {P\left( {{X_1},{X_2}|S_{_1}^i,S_{_2}^i} \right)} } \right)$. These neural activities are then fed into the third layer with synaptic weights ${w_1}$ and ${w_2}$, where ${w_1} = I\left( {S_1^i = S_2^i} \right)$ and ${w_2} = I\left( {S_1^i \ne S_2^i} \right)$. In the fourth layer, a max operation is taken to decide whether the cause is 1 or 2. Note that, the precondition of the inference is that we have known the prior probability and conditional probability. We suppose that the prior probability is presented by the distribution of Poisson spiking neurons, which means the states of the Poisson spiking neurons follow the prior distribution. We also assume that the tuning curves are proportional to conditional probability. With the benefit of sampling-based inference, massive number of neurons could sample in parallel and calculate without iteration. This means that the neural circuit could trade space for time thus the inference would be quite rapid.

\section{Experiments}
In this section, we demonstrate the merits of the proposed method with experiments. Samples of ${X_1}$ and ${X_2}$ are generated according to the prior probabilities $P\left( C \right)$ and conditional probabilities $P\left( {{S_1},{S_2}|C} \right)$, $P\left( {{X_1}|{S_1}} \right)$,
$P\left( {{X_2}|{S_2}} \right)$ in three steps. First, we generate samples of variables $C$ with equality probabilities 0.5 for each state. Then for each sample ${C^i}$, if ${C^i} = 1$, $S_1^i$ will be generated from a Gaussian distribution with mean 0 and variances $\sigma _S^2$, and $S_2^i = S_1^i$. If ${C^i} = 2$, $S_1^i$ and $S_2^i$ will be drawn from the same Gaussian distribution whose mean is 0 and variances is $\sigma _S^2$. At last $X_1^i$ and $X_2^i$ are generated from two different Gaussian distributions, whose means are $S_1^i$ and $S_2^i$ and variances are $\sigma _1^2$ and $\sigma _2^2$ respectively.

\subsection{Experiment 1: Simulating the Poisson spiking neurons and their firing rates }
\begin{figure}
\centering
\includegraphics{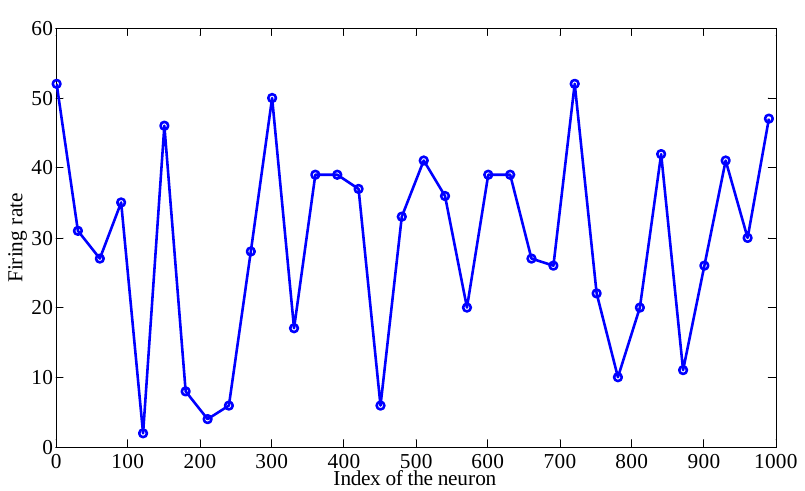}
\caption{The firing rate of the Poisson spiking neurons in a trial.}
\label{fig_5}
\end{figure}
\begin{figure}
\centering
\includegraphics{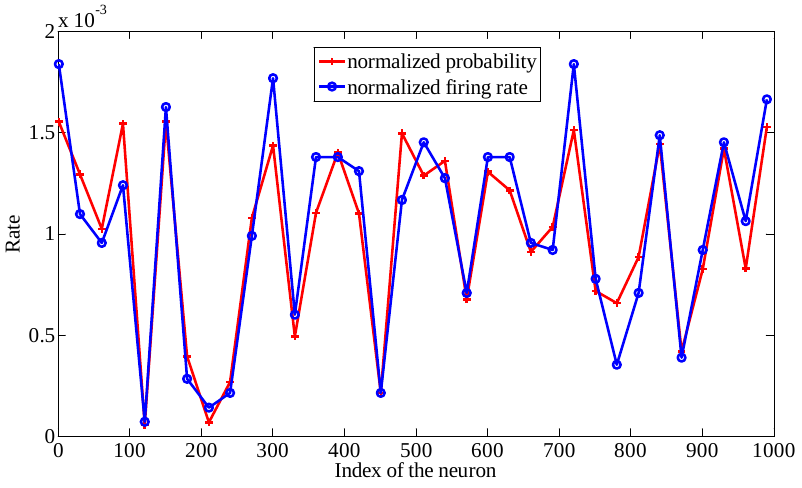}
\caption{The normalized firing rate compared with the normalized distribution in a trial.}
\label{fig_6}
\end{figure}
Here we simulate the behaviors of the Poisson spiking neurons in Fig. \ref{fig_4}. We first generate the input $X_1^i$ and $X_2^i$ randomly with the method proposed above. The parameters are specific to ${\sigma _S} = 4$, ${\sigma _1} = {\sigma _2} = 6$. Then we generate 1000 Poisson spiking neurons $S_1^1S_2^1$, $S_1^2S_2^2$,..., $S_1^{1000}S_2^{1000}$ by sampling from $P\left( {{S_1},{S_2}} \right)$. The tuning curve of the neuron $S_1^iS_2^i$ is set to $10000 \times {P\left( {{X_1},{X_2}|S_1^i,S_2^i} \right)}$. Fig. \ref{fig_5} represents the firing rate $r_1^1,r_1^2,...,r_1^{1000}$ of the 1000 Poisson spiking neurons in a trial (the firing rate could vary in different trials due to neural variability). Note that here we only show the results of neurons whose indexes range from 1 to 1000 with a uniform spacing 30. In Fig. \ref{fig_6}, the circles on the blue curve represents the normalized firing rate for neuron $S_1^iS_2^i$, which is $r_1^i/\sum\limits_{i = 1}^{1000} {r_1^i} $. Similarly, we only show the results of neurons whose indexes range from 1 to 1000 with a uniform spacing 30. The plus on the red curve is the normalized probability $P\left( {{X_1},{X_2}|S_1^i,S_2^i} \right)/\sum\limits_{i = 1}^{1000} {P\left( {{X_1},{X_2}|S_1^i,S_2^i} \right)} $. We can see that the normalized firing rate is close to the normalized probability.

\subsection{Experiment 2: Testing on the convergence and accuracy of our method }

Here we simulate and present the behaviors of the neurons in the last two layers and show the convergence and accuracy of our method. We first generate 1000 inputs of $X_1^i$ and $X_2^i$ randomly with the method proposed above. For each inputs $X_1^i$ and $X_2^i$,  ${\sigma _S}$, ${\sigma _1}$ and ${\sigma _2}$ are drawn randomly from a uniform distribution on [3 7]. Then we calculate $P\left( {C = 1|X_1^i,X_2^i} \right)$ with the sampling-based method and express the result as ${P^{{\rm{sample}}}}\left( {C = 1|X_1^i,X_2^i} \right)$. Meanwhile, the truth of $P\left( {C = 1|X_1^i,X_2^i} \right)$ is expressed as ${P^{{\rm{truth}}}}\left( {C = 1|X_1^i,X_2^i} \right)$, which is calculated with the elimination method in [16]. The error of samples $X_1^i$ and $X_2^i$ is defined by $\delta  = \left| {{P^{{\rm{sample}}}}\left( {C = 1|X_1^i,X_2^i} \right) - {P^{{\rm{truth}}}}\left( {C = 1|X_1^i,X_2^i} \right)} \right|$. This index expresses the gap between the sampling value and optimal value of posterior probability. The mean error is calculated from 1000 different inputs. The error rate represents the proportion of false results in the 1000 different inputs when we infer the cause.
\begin{figure}
\centering
\includegraphics{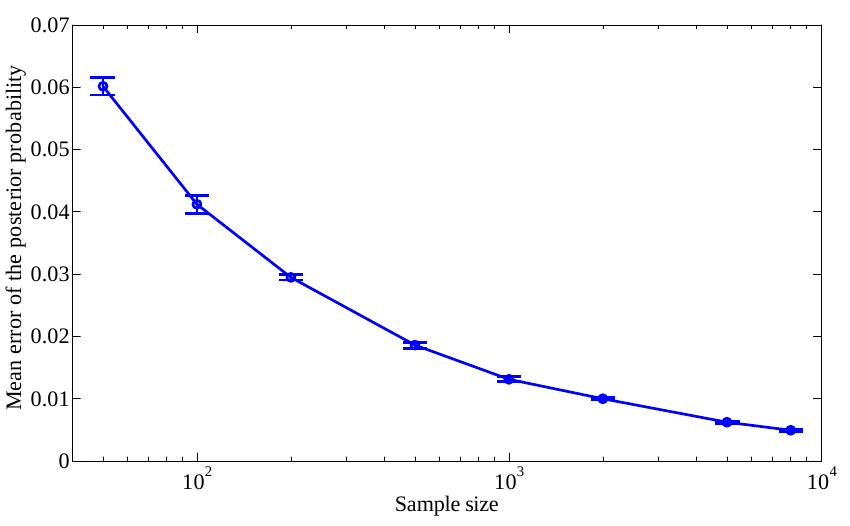}
\caption{Mean error of the posterior probability varies with sample size.}
\label{fig_7}
\end{figure}
\begin{figure}
\centering
\includegraphics{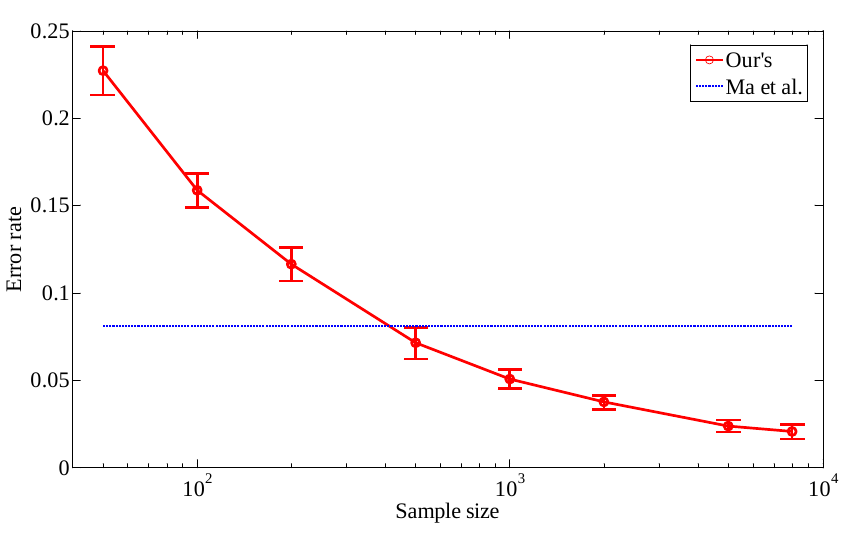}
\caption{Error rate of causal inference.}
\label{fig_8}
\end{figure}
We calculate the mean error and the error rate with different sample sizes and repeat the experiments 10 times. The mean error of the posterior probability varying with sample size is shown in Fig. \ref{fig_7}. We find that the mean error decreases as the sample size increases and converges to zero when the sample size tends to infinity. This result demonstrates that if we have enough samples, or we have enough neurons, our algorithm will get the optimal value. Fig. \ref{fig_8} plots the error rate obtained from our method and that from the method of Ma et al [16]. The method of Ma et al. is not related to sample size while ours could get stretchable results (different accuracies) with different sample sizes. Obviously, our method is superior to Ma's when the sample size is larger than 500 (500 neurons for each variable).  We can see that in order to keep the error rate under 0.05 for two-stimuli causal inference, we need at least 1000 neurons to represent each variable. This means $N = 1000$ in equation (3).

\subsection{Experiment 3: Testing on the applicability of the method with different parameters. }
\begin{figure}
\centering
\includegraphics{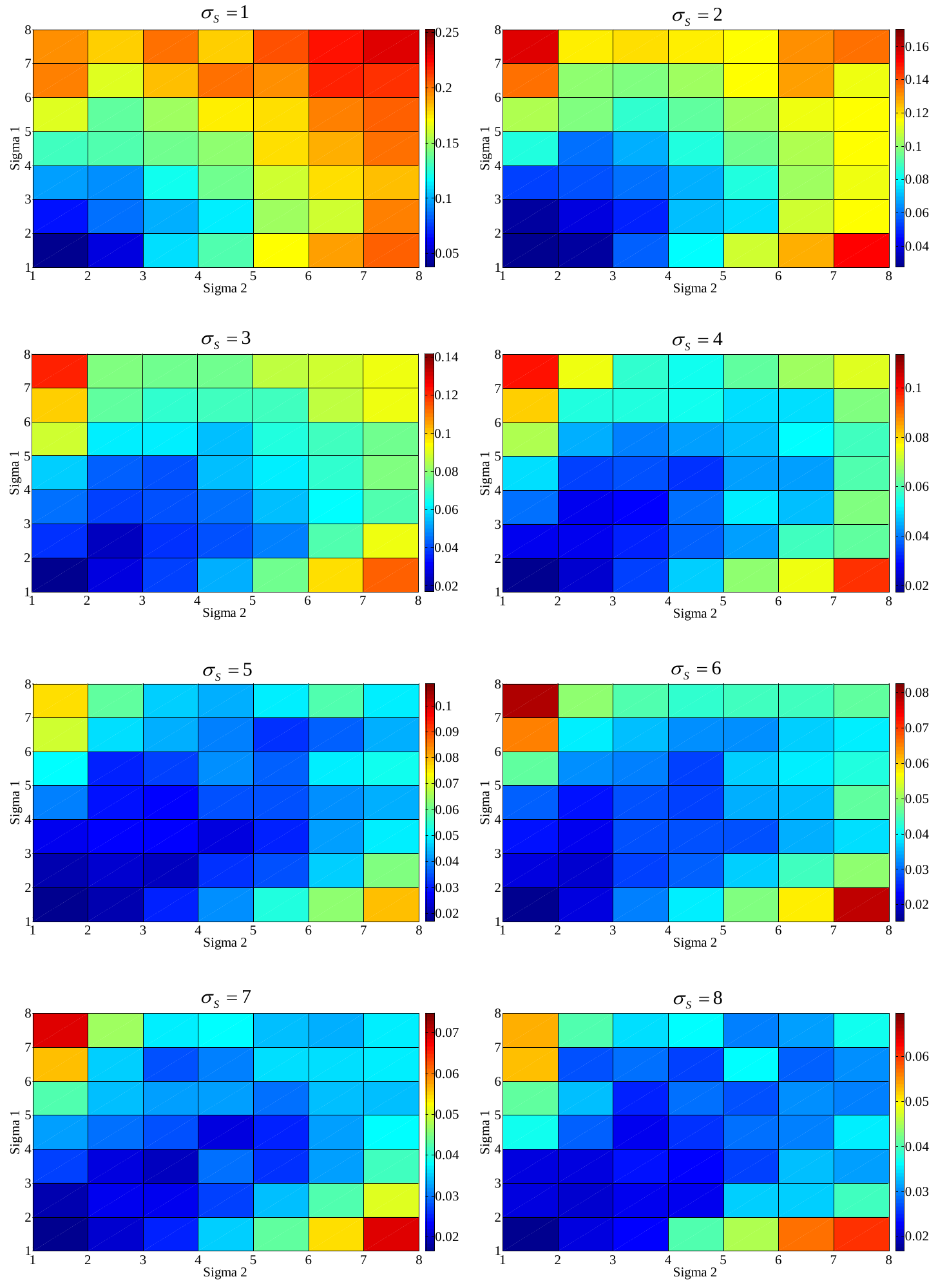}
\caption{Error rate for different parameters with sample size equals 1000.}
\label{fig_9}
\end{figure}

Experiment 1 and 2 indicate that our inference method could get the optimal solution given enough samples. However, the parameters ${\sigma _1}$, ${\sigma _2}$ and ${\sigma _S}$ are drawn randomly for each trial. In experiment 3, we will make a concrete analysis of the applicability of our method with different parameters. In this experiment, ${\sigma _1}$,  ${\sigma _2}$ and ${\sigma _S}$ can vary from 1 to 8. We test the error rate for different ${\sigma _1}$, ${\sigma _2}$ and ${\sigma _S}$ with the sample size being 1000 and show the result in Fig. \ref{fig_9}. In each sub-figure, ${\sigma _S}$ is set to a fixed value while both ${\sigma _2}$ and ${\sigma _S}$ vary from 1 to 8. We can see that the error rate is less than 0.1 for most of the parameters. However, when ${\sigma _S}$ is in close proximity to zero, the error rate turns out to be very high. This could be explained as follows. Since ${\sigma _S}$ is close to zero, the difference between ${S_1}$ and ${S_2}$ remains very small no matter $C = 1\;or\;2$. Then the posterior probability of a common cause and two different causes both will be quite near to 0.5. Due to this, a very small error could lead to incorrect inference results, making the error rate very high. Nevertheless, if there are adequate samples, the error rate could be arbitrarily small, which means our method is robust to different parameters.

\subsection{Experiment 4: Testing on the probability of reporting a common cause with respect to stimulus disparity. }
\begin{figure}
\centering
\includegraphics{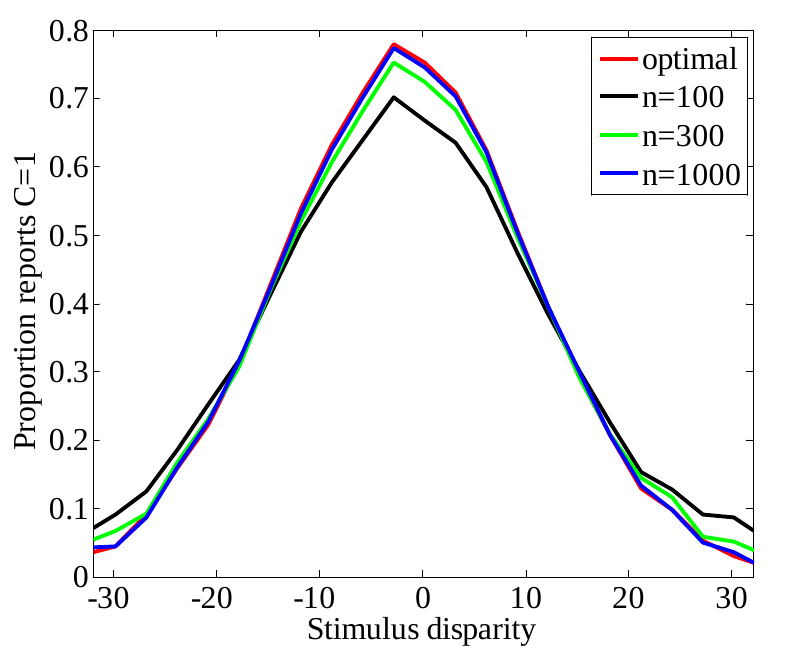}
\caption{Comparison of proportion reports ${C} = 1$ with respect to stimulus disparity between accurate equation and our sampling-based method.}
\label{fig_10}
\end{figure}
Stimulus disparity refers to the space difference between different stimuli, which is defined by ${S_2} - {S_1}$, where ${S_1}$ and ${S_2}$ are two different stimuli. Intuitively, it is more likely that there is a common cause if stimulus disparity is small, while there are two different causes if stimulus disparity is large. In this experiment, 200000 samples are generated with parameters ${\sigma _1} = 3$, ${\sigma _S} = {\sigma _2} = 10$. For each $S_1^i$ and $S_2^i$, stimulus disparity is defined by $S_2^i - S_1^i$. The state of variable ${C^i}$ is inferred by optimal equation \cite{ma2013towards} and our method respectively. Then for the samples with the same stimulus disparity, we calculate the proportion of reporting a common cause. Fig. \ref{fig_10} shows the result, the red curve is obtained by optimal equation. The black, green and blue curves are calculated by our method with sample size being 100, 300 and 1000 respectively. The result shows that as the sample size becomes larger, the sampling-based curve tends to be closer to the accuracy curve. When sample size equals 1000, the sampling-based curve is almost the same as the accuracy curve. This result indicates the accuracy of our method.

\section{Generalization}
In this section, we generalize our sampling-based method to implement inference for other two important problems: the multi-stimuli causal inference in cue combination and the same-different judgment.
\subsection{Multi-stimuli causal inference}
\begin{figure}
\centering
\includegraphics{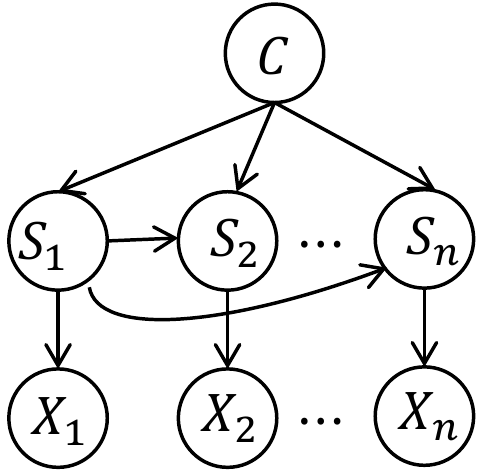}
\caption{The Bayesian model for the multi-stimuli causal inference.}
\label{fig_11}
\end{figure}

In the above experiments, situations where there are only two stimuli were taken into consideration while in our daily life cues may come from multiple sensory modalities, such as visual, auditory, and tactile. Despite the fact that Bayesian model could also explain the causal inference problem with multi-stimuli \cite{wozny2008human}, how to implement inference with neural circuits is unclear. With our sampling-based method, it is easy to generalize the neural circuits to implement inference for multi-stimuli.

The steps are similar to that in section 3. First we convert the problem to a Bayesian network (shown in Fig. \ref{fig_11}). The prior probabilities and conditional probabilities are defined according to the causal inference model, where different states of $C$ reflect different situations of the cause. The causal inference problem is then converted to the inference of posterior probability, which could be calculated by importance sampling:

\begin{equation}
\label{eq_5}
\begin{array}{l}
\;P\left( {C = 1|{X_1} = {x_1},{X_2} = {x_2},...,{X_n} = {x_n}} \right)\\
 = \;P\left( {C = 1|{x_1},{x_2},...,{x_n}} \right)\\
 = \int\limits_{{S_1},{S_2},...,{S_n}} {P\left( {C = 1,{S_1},{S_2},...,{S_n}|{x_1},{x_2},...,{x_n}} \right)d{S_1},{S_2},...,{S_n}} \\
 = \int\limits_{{S_1},{S_2},...,{S_n}} {P\left( {C = 1|{S_1},{S_2},...,{S_n}} \right)P\left( {{S_1},{S_2},...,{S_n}|{x_1},{x_2},...,{x_n}} \right)d{S_1},{S_2},...,{S_n}} \\
 = \frac{{\int\limits_{{S_1},{S_2},...,{S_n}} {P\left( {C = 1|{S_1},{S_2},...,{S_n}} \right)P\left( {{x_1},{x_2},...,{x_n}|{S_1},{S_2},...,{S_n}} \right)P\left( {{S_1},{S_2},...,{S_n}} \right)d{S_1},{S_2},...,{S_n}} }}{{\int\limits_{{S_1},{S_2},...,{S_n}} {P\left( {{x_1},{x_2},...,{x_n}|{S_1},{S_2},...,{S_n}} \right)P\left( {{S_1},{S_2},...,{S_n}} \right)d{S_1},{S_2},...,{S_n}} }}\\
 = \frac{{E{{\left( {P\left( {C = 1|{S_1},{S_2},...,{S_n}} \right)P\left( {{x_1},{x_2},...,{x_n}|{S_1},{S_2},...,{S_n}} \right)} \right)}_{P\left( {{S_1},{S_2},...,{S_n}} \right)}}}}{{E{{\left( {P\left( {{x_1},{x_2},...,{x_n}|{S_1},{S_2},...,{S_n}} \right)} \right)}_{P\left( {{S_1},{S_2},...,{S_n}} \right)}}}}\\
 \approx \sum\limits_{\scriptstyle i = 1\atop
\scriptstyle S_{_1}^i,...,S_{_n}^i \sim P\left( {{S_1},...,{S_n}} \right)}^N {P\left( {C = 1|S_{_1}^i,S_{_2}^i,...,S_{_n}^i} \right)} \frac{{P\left( {{x_1},...,{x_n}|S_{_1}^i,S_{_2}^i,...,S_{_n}^i} \right)}}{{\sum\limits_{\scriptstyle i = 1\atop
\scriptstyle S_{_1}^i,...,S_{_n}^i \sim P\left( {{S_1},...,{S_n}} \right)}^N {P\left( {{x_1},...,{x_n}|S_{_1}^i,S_{_2}^i,...,S_{_n}^i} \right)} }}\\
 = \sum\limits_{\scriptstyle i = 1\atop
\scriptstyle S_{_1}^i,...,S_{_n}^i \sim P\left( {{S_1},...,{S_n}} \right)}^N {\frac{{P\left( {S_{_1}^i,...,S_{_n}^i|C = 1} \right)}}{{P\left( {S_{_1}^i,...,S_{_n}^i|C = 1} \right) + P\left( {S_{_1}^i,...,S_{_n}^i|C = 2} \right)}}} \frac{{P\left( {{x_1},...,{x_n}|S_{_1}^i,...,S_{_n}^i} \right)}}{{\sum\limits_{\scriptstyle i = 1\atop
\scriptstyle S_{_1}^i,...,S_{_n}^i \sim P\left( {{S_1},...,{S_n}} \right)}^N {P\left( {{x_1},...,{x_n}|S_{_1}^i,...,S_{_n}^i} \right)} }}\\
 = \sum\limits_{\scriptstyle i = 1\atop
\scriptstyle S_{_1}^i,...,S_{_n}^i \sim P\left( {{S_1},...,{S_n}} \right)}^N {I\left( {S_{_1}^i = S_{_2}^i = ... = S_{_n}^i} \right)\frac{{P\left( {{x_1},...,{x_n}|S_{_1}^i,S_{_2}^i,...,S_{_n}^i} \right)}}{{\sum\limits_{\scriptstyle i = 1\atop
\scriptstyle S_{_1}^i,...,S_{_n}^i \sim P\left( {{S_1},...,{S_n}} \right)}^N {P\left( {{x_1},...,{x_n}|S_{_1}^i,S_{_2}^i,...,S_{_n}^i} \right)} }}}
\end{array}
\end{equation}

Equation (\ref{eq_5}) is similar to (\ref{eq_3}) except that the stimuli here are ${S_1},{S_2},...,{S_n}$ rather than ${S_1},{S_2}$. According to (\ref{eq_5}), the neural circuit of multi-stimuli
causal inference is similar to that of two-stimuli causal inference except three differences. Firstly, the states of the Poission spiking neurons sample from $P\left( {{S_1},{S_2},...,{S_n}} \right)$, rather than $P\left( {{S_1},{S_2}} \right)$. Secondly, the tuning curve of the neuron marked as $i$ is proportional to ${P\left( {{x_1},{x_2},...,{x_n}|S_{_1}^i,S_{_2}^i,...,S_n^i} \right)}$. Thirdly, the synaptic weights are ${I\left( {S_{_1}^i = S_{_2}^i = ... = S_{_n}^i} \right)}$ instead of ${I\left( {S_{_1}^i = S_{_2}^i} \right)}$.

\begin{figure}
\centering
\includegraphics{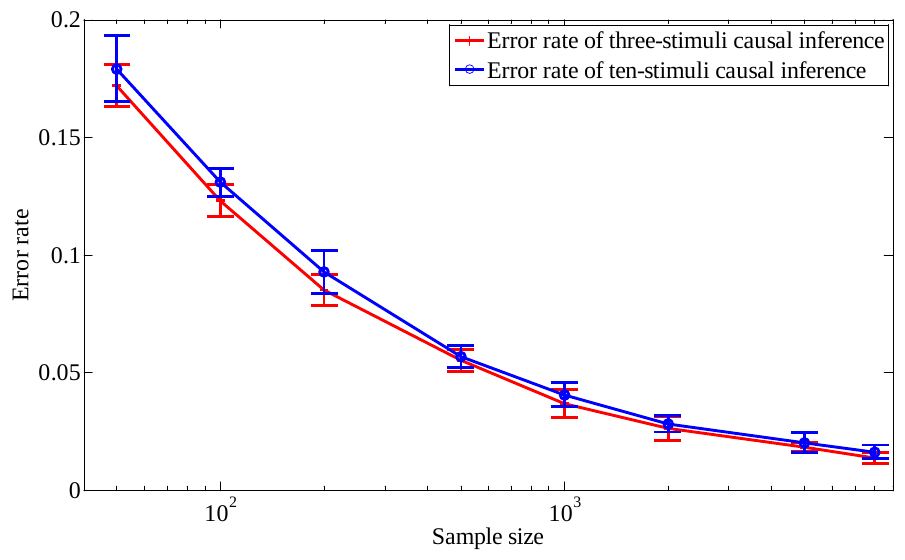}
\caption{Error rate of the multi-stimuli causal inference.}
\label{fig_12}
\end{figure}

We also test on the convergence and accuracy of our method for multi-stimuli causal inference to show its feasibility. Three-stimuli casual inference is tested because the situation is quite common in our daily life, such as the integration of visual, auditory, and tactile input. We also test ten-stimuli to show that our method applies to higher dimensions. We assume that $P\left( {C = 1} \right)$ is equal to $P\left( {C = 2} \right)$, both of which have a probability 0.5. Note that $C = 1$ means the three cues have the same cause and $C = 2$ means the cues have diverse causes. We define $P\left( {{S_1},{S_2},...,{S_t}|C = 1} \right) = \frac{1}{{\sqrt {2\pi } {\sigma _S}}}\exp \left( { - \frac{{S_1^2}}{{2\sigma _S^2}}} \right)\prod\limits_{j = 1}^t {\delta \left( {{S_1} - {S_j}} \right)} $ and $P\left( {{S_1},{S_2},...,{S_t}|C = 2} \right) = \frac{1}{{{{\left( {\sqrt {2\pi \sigma _S^2} } \right)}^t}}}\;\;$  $\exp \left( { - \frac{{S_1^2 + S_2^2 + ... + S_t^2}}{{2\sigma _S^2}}} \right)$. Besides, the conditional probability of ${X_i}$ under ${S_i}\;\;$ $\left( {i = 1,2,...,t} \right)$ is defined by $P\left( {{X_i}|{S_i}} \right) = \frac{1}{{\sqrt {2\pi } {\sigma _i}}}\exp \left( { - \frac{{{{\left( {{X_i} - {S_i}} \right)}^2}}}{{2\sigma _i^2}}} \right)$. The experimental procedure is similar to that of Experiment 2 in section 3 and the result is shown in Fig. \ref{fig_12}. We can find that the error rate decreases as the sample size increases and convergences to zero when sample size tends to infinity. We also find that we don't need to scale up the samples when dimensions become higher. 1000 samples (neurons) are required for each variable to keep the error rate under 0.05. These results are in good agreement with the fact that importance sampling does not scale up with higher dimensions.

\subsection{ Same-different judgment}

When faced with multiple objects, probably the first thing our brain needs to do is to decide whether they are the same or not. Thus the same-different judgment could be critical in perception and cognition. A straightforward example is object classification. Human brains are able to recognize the same object and assign them to the semantic classes. Berg et al. \cite{van2012optimal} propose the optimal-observer model and prove that the same-different judgment is a process of probability inference. As illustrated in Fig. \ref{fig_13}, variable $C$ represents the judgment, $C = 1$ means the objects are the same while $C = 2$ means they are different. $\mu $ is a single value parameter variable generated from a uniform distribution ranging from $-L$ to $L$.  When the  objects are the same, ${S_i}$ equals to $\mu $. When they are different, ${S_i}$ is drawn from a Gaussian distribution with mean ${\mu _i}$ and variance $\sigma _S^2$.  The distribution of ${X_i}$ is a Gaussian distribution with its mean being ${S_i}$ and its variance being $\sigma _i^2$. Based on these definitions, the same-different judgment problem can be converted to the posterior probability inference problem of variable $C$, which could be calculated by importance sampling:
\begin{figure}
\centering
\includegraphics{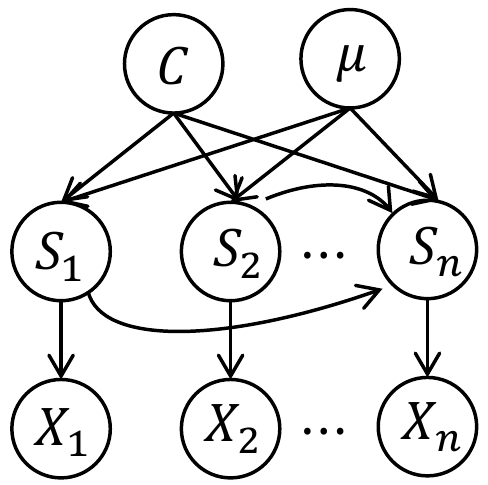}
\caption{Optimal-observer model of the same-different judgment problem.}
\label{fig_13}
\end{figure}

\begin{equation}
\label{eq_6}
\begin{array}{l}
\;P\left( {C = 1|{X_1} = {x_1},{X_2} = {x_2},...,{X_n} = {x_n}} \right)\\
 = \;P\left( {C = 1|{x_1},{x_2},...,{x_n}} \right)\\
 = \int\limits_{{S_1},...,{S_n}} {P\left( {C = 1,{S_1},...,{S_n}|{x_1},...,{x_n}} \right)d{S_1},...,{S_n}} \\
 = \int\limits_{{S_1},...,{S_n}} {P\left( {C = 1|{S_1},...,{S_n}} \right)P\left( {{S_1},...,{S_n}|{x_1},...,{x_n}} \right)d{S_1},...,{S_n}} \\
 \approx \sum\limits_{\scriptstyle i = 1\atop
\scriptstyle S_{_1}^i,...,S_{_n}^i \sim P\left( {{S_1},...,{S_n}} \right)}^N {P\left( {C = 1|S_{_1}^i,...,S_{_n}^i} \right)\frac{{P\left( {{x_1},...,{x_n}|S_{_1}^i,...,S_{_n}^i} \right)}}{{\sum\limits_{\scriptstyle i = 1\atop
\scriptstyle S_{_1}^i,...,S_{_n}^i \sim P\left( {{S_1},...,{S_n}} \right)}^N {P\left( {{x_1},...,{x_n}|S_{_1}^i,...,S_{_n}^i} \right)} }}} \\
 = \sum\limits_{\scriptstyle i = 1\atop
\scriptstyle S_{_1}^i,...,S_{_n}^i \sim P\left( {{S_1},...,{S_n}} \right)}^N {\frac{{P\left( {S_{_1}^i,...,S_{_n}^i|C = 1} \right)}}{{P\left( {S_{_1}^i,...,S_{_n}^i|C = 1} \right) + P\left( {S_{_1}^i,...,S_{_n}^i|C = 2} \right)}}\frac{{P\left( {{x_1},...,{x_n}|S_{_1}^i,...,S_{_n}^i} \right)}}{{\sum\limits_{\scriptstyle i = 1\atop
\scriptstyle S_{_1}^i,...,S_{_n}^i \sim P\left( {{S_1},...,{S_n}} \right)}^N {P\left( {{x_1},...,{x_n}|S_{_1}^i,...,S_{_n}^i} \right)} }}} \;\;\;\;\;\;\;\;\;\;\;\\
 = \sum\limits_{\scriptstyle i = 1\atop
\scriptstyle S_{_1}^i,...,S_{_n}^i \sim P\left( {{S_1},...,{S_n}} \right)}^N {\left( {\frac{{\int\limits_\mu  {P\left( {S_{_1}^i,...,S_{_n}^i,\mu |C = 1} \right)d\mu } }}{{\int\limits_\mu  {P\left( {S_{_1}^i,...,S_{_n}^i,\mu |C = 1} \right)d\mu }  + \int\limits_\mu  {P\left( {S_{_1}^i,...,S_{_n}^i,\mu |C = 2} \right)d\mu } }}} \right.} \\
\;\;\;\\
\;\;\;\;\;\;\;\;\;\;\;\;\;\;\left. {\frac{{P\left( {{x_1},...,{x_n}|S_{_1}^i,...,S_{_n}^i} \right)}}{{\sum\limits_{\scriptstyle i = 1\atop
\scriptstyle S_{_1}^i,...,S_{_n}^i \sim P\left( {{S_1},...,{S_n}} \right)}^N {P\left( {{x_1},...,{x_n}|S_{_1}^i,...,S_{_n}^i} \right)} }}} \right)\\
 = \sum\limits_{\scriptstyle i = 1\atop
\scriptstyle S_{_1}^i,...,S_{_n}^i \sim P\left( {{S_1},...,{S_n}} \right)}^N {\left( {\;\frac{{\int\limits_\mu  {P\left( {S_{_1}^i,...,S_{_n}^i|C = 1,\mu } \right)d\mu } }}{{\int\limits_\mu  {P\left( {S_{_1}^i,...,S_{_n}^i|C = 1,\mu } \right)d\mu }  + \int\limits_\mu  {P\left( {S_{_1}^i,...,S_{_n}^i|C = 2,\mu } \right)d\mu } }}} \right.} \\
\;\;\;\;\;\;\;\;\;\;\;\;\;\;\left. {\frac{{P\left( {{x_1},...,{x_n}|S_{_1}^i,...,S_{_n}^i} \right)}}{{\sum\limits_{\scriptstyle i = 1\atop
\scriptstyle S_{_1}^i,...,S_{_n}^i \sim P\left( {{S_1},...,{S_n}} \right)}^N {P\left( {{x_1},...,{x_n}|S_{_1}^i,...,S_{_n}^i} \right)} }}} \right)\\
 = \sum\limits_{\scriptstyle i = 1\atop
\scriptstyle S_{_1}^i,...,S_{_n}^i \sim P\left( {{S_1},...,{S_n}} \right)}^N {I\left( { - L \le S_{_1}^i = S_{_2}^i = ... = S_{_n}^i \le L} \right)\frac{{P\left( {{x_1},...,{x_n}|S_{_1}^i,...,S_{_n}^i} \right)}}{{\sum\limits_{\scriptstyle i = 1\atop
\scriptstyle S_{_1}^i,...,S_{_n}^i \sim P\left( {{S_1},...,{S_n}} \right)}^N {P\left( {{x_1},...,{x_n}|S_{_1}^i,...,S_{_n}^i} \right)} }}}
\end{array}
\end{equation}

In equation (\ref{eq_6}), ${I\left( { - L \le S_{_1}^i = S_{_2}^i = ... = S_{_n}^i \le L} \right)}$ is a indicative function. It equals to 1 only when all the stimuli are the same and between $-L$ and $L$. Equation (\ref{eq_6}) differs from equation (\ref{eq_5}) in the indicative function. Due to this, the neural circuit for the same-different judgment are similar to that for multi-stimuli causal inference except one difference. That is, the synaptic weights of the neural circuit for the same-different judgment are ${I\left( { - L \le S_{_1}^i = S_{_2}^i = ... = S_{_n}^i \le L} \right)}$, instead of
${I\left( {S_{_1}^i = S_{_2}^i = ... = S_{_n}^i} \right)}$.

\begin{figure}
\centering
\includegraphics{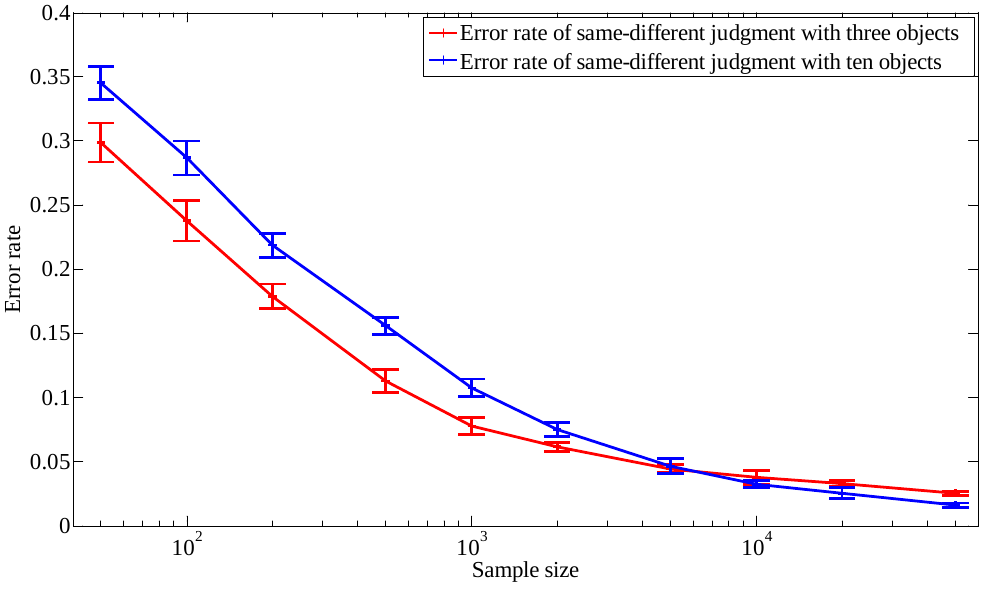}
\caption{Error rate of the same-different judgment}
\label{fig_14}
\end{figure}

We test the accuracy and convergence of our method for the same-different judgment problem. In this experiment, we compute the error rate of same-different judgment with three objects and ten objects respectively. Samples of ${X_1}$, ${X_2}$...., and ${X_t}$ are generated according to the similar method in the section of multi-stimuli causal inference. Note that ${\mu _i}$ is generated from the uniform distribution ranging from $-10$ to $10$.  ${\sigma _S}$ and ${\sigma _i}$ are drawn randomly from a uniform distribution on [1 3]. The inference result with our method is present in Fig. \ref{fig_14}, which is similar to that of Fig. \ref{fig_13}. We can find that the error rate of same-different judgment with three objects and ten objects both decrease as the sample size increases and convergence to zero when sample size tends to infinity. Besides, the samples needed don't scale up when dimensions become higher. 5000 samples (neurons) are required for each variable to keep the error rate under 0.05.

\section{Conclusion}
In this paper, we propose an inference algorithm for causal inference in cue combination based on importance sampling and design a corresponding neural circuit to implement this inference algorithm. The neural circuit is plausible as it is based on PPC and three types of plausible neural operations. Theoretical analysis and experimental results show that our algorithm can converge to the accurate value as sample size goes to infinite. It is worth noting that our method provides a general solution to the other two important problems, namely the multi-stimuli causal inference and the same-different judgment.

Different from Markov chain Monte Carlo \cite{buesing2011neural,pecevski2011probabilistic}, which represents the distribution with variability of a neuron over time, our method utilizes the variability over neurons to represent the distribution. This means that massive number of neurons could sample in parallel and calculate without iteration, thus the inference would be quite rapid.

Despite the plausible neural implementation of inference, the question of how to learn the prior probabilities and conditional probabilities with learning rules found in biological studies requires considerable future work. Besides, learning and inference should be implemented by the same neural circuits. Some recent works have provided reference experiences for implementing learning. For example, Maass et al. prove that Spike-Timing-Dependent Plasticity (STDP) is able to approximate a parameter estimation algorithm--expectation maximization (EM) algorithm \cite{nessler2009stdp,nessler2013bayesian,kappel2014stdp}. This principle may be used to solve the learning problem in our paper.

\section*{Proof of Theorem 1}

\textbf{Lemma 1.} Supposing that random variables ${X^1},{X^2},...,{X^n}$ are pairwise independent and ${X^i} \sim P\left( X \right)$. Similarly, ${Y^1},{Y^2},...,{Y^n}$ are pairwise independent and ${Y^j} \sim P\left( Y \right)$. Besides, $E\left( X \right) = {\mu _1}$, $E\left( Y \right) = {\mu _2}$, ${\mu _1},{\mu _2} \ne 0$, $Var\left( X \right) = \sigma _1^2$ and $Var\left( Y \right) = \sigma _2^2$. Then for arbitrary small number $\varepsilon $, we can conclude that $P\left( {\left| {\frac{{\sum\limits_{i = 1}^N {{X^i}} }}{{\sum\limits_{j = 1}^N {{Y^j}} }} - \frac{{{\mu _1}}}{{{\mu _2}}}} \right| < \varepsilon } \right) > 1 - \frac{{16\sigma _1^2}}{{N\mu _2^2{\varepsilon ^2}}} - \frac{{16\mu _1^2\sigma _2^2}}{{N\mu _2^4{\varepsilon ^2}}}$.

\textbf{Proof:} As random variables ${X^1},{X^2},...,{X^n}$ are pairwise independent and ${X^i} \sim P\left( X \right)$, ${Y^1},{Y^2},...,{Y^n}$ are pairwise independent and ${Y^j} \sim P\left( Y \right)$, We can get
\[\begin{array}{l}
E\left( {\frac{1}{N}\sum\limits_{i = 1}^N {{X^i}} } \right) = \frac{1}{N}\sum\limits_{i = 1}^N {E\left( {{X^i}} \right)}  = {\mu _1},
\end{array}\]

\[\begin{array}{l}
E\left( {\frac{1}{N}\sum\limits_{j = 1}^N {{Y^j}} } \right) = \frac{1}{N}\sum\limits_{j = 1}^N {E\left( {{Y^j}} \right)}  = {\mu _2},
\end{array}\]

\[\begin{array}{l}
Var\left( {\frac{1}{N}\sum\limits_{i = 1}^N {{X^i}} } \right) = \frac{1}{{{N^2}}}\sum\limits_{i = 1}^N {Var\left( {{X^i}} \right)}  = \frac{{\sigma _1^2}}{N},
\end{array}\]

\[\begin{array}{l}
Var\left( {\frac{1}{N}\sum\limits_{j = 1}^N {{Y^j}} } \right) = \frac{1}{{{N^2}}}\sum\limits_{j = 1}^N {Var\left( {{Y^j}} \right)}  = \frac{{\sigma _2^2}}{N}.
\end{array}\]
For arbitrary small number ${\varepsilon _1}$ and ${\varepsilon _2} = \frac{{{\mu _2}}}{{{\mu _1}}}{\varepsilon _1}$, application of the Chebyshev's Inequality yields the inequality:

\[\begin{array}{l}
\;P\left( {\left| {\frac{1}{N}\sum\limits_{i = 1}^N {{X^i} - {\mu _1}} } \right| < {\varepsilon _1}} \right) \ge 1 - \frac{{\sigma _1^2}}{{N\varepsilon _1^2}},
\end{array}\]

\[\begin{array}{l}
P\left( {\left| {\frac{1}{N}\sum\limits_{j = 1}^N {{Y^j} - {\mu _2}} } \right| < {\varepsilon _2}} \right) \ge 1 - \frac{{\sigma _2^2}}{{N\varepsilon _2^2}},
\end{array}\]
which is equivalent to

\[\begin{array}{l}
P\left( {{\mu _1} - {\varepsilon _1} < \frac{1}{N}\sum\limits_{i = 1}^N {{X^i}}  < {\mu _1} + {\varepsilon _1}} \right) \ge 1 - \frac{{\sigma _1^2}}{{N\varepsilon _1^2}},
\end{array}\]

\[\begin{array}{l}
P\left( {{\mu _2} - {\varepsilon _2} < \frac{1}{N}\sum\limits_{j = 1}^N {{Y^j}}  < {\mu _2} + {\varepsilon _2}} \right) \ge 1 - \frac{{\sigma _2^2}}{{N\varepsilon _2^2}}.
\end{array}\]
Because $\frac{1}{N}\sum\limits_{i = 1}^N {{X^i}} $ and $\frac{1}{N}\sum\limits_{j = 1}^N {{Y^j}} $ are independent, so it is trivial to show that

\[\begin{array}{l}
P\left( {\frac{{{\mu _1} - {\varepsilon _1}}}{{{\mu _2} + {\varepsilon _2}}} < \frac{{\sum\limits_{i = 1}^N {{X^i}} }}{{\sum\limits_{j = 1}^N {{Y^j}} }} < \frac{{{\mu _1} + {\varepsilon _1}}}{{{\mu _2} - {\varepsilon _2}}}} \right) \ge \left( {1 - \frac{{\sigma _1^2}}{{N\varepsilon _1^2}}} \right)\left( {1 - \frac{{\sigma _2^2}}{{N\varepsilon _2^2}}} \right).
\end{array}\]
An application of Taylor's formula yields:

\[\begin{array}{l}
\frac{{{\mu _1} - {\varepsilon _1}}}{{{\mu _2} + {\varepsilon _2}}} = \frac{{{\mu _1}}}{{{\mu _2}}}\frac{{1 - \frac{{{\varepsilon _1}}}{{{\mu _1}}}}}{{1 + \frac{{{\varepsilon _1}}}{{{\mu _1}}}}} = \frac{{{\mu _1}}}{{{\mu _2}}}\left( {1 - \frac{{2\frac{{{\varepsilon _1}}}{{{\mu _1}}}}}{{1 + \frac{{{\varepsilon _1}}}{{{\mu _1}}}}}} \right)\\
\;\;\;\;\;\;\;\;\;\;\; = \frac{{{\mu _1}}}{{{\mu _2}}} - \frac{{{\mu _1}}}{{{\mu _2}}}\frac{{2{\varepsilon _1}}}{{{\mu _1}}}\left( {1 - \frac{{{\varepsilon _1}}}{{{\mu _1}}} + {{\left( {\frac{{{\varepsilon _1}}}{{{\mu _1}}}} \right)}^2} - {{\left( {\frac{{{\varepsilon _1}}}{{{\mu _1}}}} \right)}^3} + ...} \right)\\
\;\;\;\;\;\;\;\;\;\;\; > \frac{{{\mu _1}}}{{{\mu _2}}} - \frac{{2{\varepsilon _1}}}{{{\mu _2}}} > \frac{{{\mu _1}}}{{{\mu _2}}} - \frac{{4{\varepsilon _1}}}{{{\mu _2}}},
\end{array}\]

\[\begin{array}{l}
\frac{{{\mu _1} + {\varepsilon _1}}}{{{\mu _2} - {\varepsilon _2}}} = \frac{{{\mu _1}}}{{{\mu _2}}}\frac{{1 + \frac{{{\varepsilon _1}}}{{{\mu _1}}}}}{{1 - \frac{{{\varepsilon _1}}}{{{\mu _1}}}}} = \frac{{{\mu _1}}}{{{\mu _2}}}\left( {1 + \frac{{2\frac{{{\varepsilon _1}}}{{{\mu _1}}}}}{{1 - \frac{{{\varepsilon _1}}}{{{\mu _1}}}}}} \right)\\
\;\;\;\;\;\;\;\;\;\;\; = \frac{{{\mu _1}}}{{{\mu _2}}} + \frac{{{\mu _1}}}{{{\mu _2}}}\frac{{2{\varepsilon _1}}}{{{\mu _1}}}\left( {1 + \frac{{{\varepsilon _1}}}{{{\mu _1}}} + {{\left( {\frac{{{\varepsilon _1}}}{{{\mu _1}}}} \right)}^2} + {{\left( {\frac{{{\varepsilon _1}}}{{{\mu _1}}}} \right)}^3} + ...} \right)\\
\;\;\;\;\;\;\;\;\;\;\; < \frac{{{\mu _1}}}{{{\mu _2}}} + \frac{{{\mu _1}}}{{{\mu _2}}}\frac{{4{\varepsilon _1}}}{{{\mu _1}}} = \frac{{{\mu _1}}}{{{\mu _2}}} + \frac{{4{\varepsilon _1}}}{{{\mu _2}}}.
\end{array}\]
The equations above indicate that

\[\begin{array}{l}
P\left( {\frac{{{\mu _1}}}{{{\mu _2}}} - \frac{{4{\varepsilon _1}}}{{{\mu _2}}} < \frac{{\sum\limits_{i = 1}^N {{X^i}} }}{{\sum\limits_{j = 1}^N {{Y^j}} }} < \frac{{{\mu _1}}}{{{\mu _2}}} + \frac{{4{\varepsilon _1}}}{{{\mu _2}}}} \right) \ge \left( {1 - \frac{{\sigma _1^2}}{{N\varepsilon _1^2}}} \right)\left( {1 - \frac{{\sigma _2^2}}{{N\varepsilon _2^2}}} \right).
\end{array}\]
Next, we rewrite the equation above as

\[\begin{array}{l}
P\left( {\left| {\frac{{\sum\limits_{i = 1}^N {{X^i}} }}{{\sum\limits_{j = 1}^N {{Y^j}} }} - \frac{{{\mu _1}}}{{{\mu _2}}}} \right| < \frac{{4{\varepsilon _1}}}{{{\mu _2}}}} \right) \ge 1 - \frac{{\sigma _1^2}}{{N\varepsilon _1^2}} - \frac{{\mu _1^2\sigma _2^2}}{{N\mu _2^2\varepsilon _1^2}} + \frac{{\mu _1^2\sigma _1^2\sigma _2^2}}{{{N^2}\mu _2^2\varepsilon _1^4}}.
\end{array}\]
Then for arbitrary small number $\varepsilon $, we have

\[\begin{array}{l}
P\left( {\left| {\frac{{\sum\limits_{i = 1}^N {{X^i}} }}{{\sum\limits_{j = 1}^N {{Y^j}} }} - \frac{{{\mu _1}}}{{{\mu _2}}}} \right| < \varepsilon } \right) \ge 1 - \frac{{16\sigma _1^2}}{{N\mu _2^2{\varepsilon ^2}}} - \frac{{16\mu _1^2\sigma _2^2}}{{N\mu _2^4{\varepsilon ^2}}} + \frac{{256\mu _1^2\sigma _1^2\sigma _2^2}}{{{N^2}\mu _2^6{\varepsilon ^4}}}.
\end{array}\]
We conclude that for arbitrary small number $\varepsilon $,

\[\begin{array}{l}
P\left( {\left| {\frac{{\sum\limits_{i = 1}^N {{X^i}} }}{{\sum\limits_{j = 1}^N {{Y^j}} }} - \frac{{{\mu _1}}}{{{\mu _2}}}} \right| < \varepsilon } \right) > 1 - \frac{{16\sigma _1^2}}{{N\mu _2^2{\varepsilon ^2}}} - \frac{{16\mu _1^2\sigma _2^2}}{{N\mu _2^4{\varepsilon ^2}}}.
\end{array}\]

\textbf{Lemma 2.} Supposing that random variables ${X^1},{X^2},...,{X^n}$ are independent pairwise and ${X^i} \sim P\left( X \right)$.Besides, we also know that $E\left( X \right) = {\mu _1}$, ${\mu _1} \ne 0$, $Var\left( X \right) = \sigma _1^2$.  Then for arbitrary small number $\varepsilon $,

\[\begin{array}{l}
P\left( {\left| {\frac{1}{{\frac{1}{N}\sum\limits_{i = 1}^N {{X^i}} }} - \frac{1}{{{\mu _1}}}} \right| < \varepsilon } \right) \ge 1 - \frac{{\sigma _1^2}}{{N\mu _1^2{\varepsilon ^2}}}.
\end{array}\]

\textbf{Proof:} The proof is similar to that of Lemma 1.

\textbf{Theorem 1.} The distributions $P\left( C \right)$, $P\left( {{S_1},{S_2}|C} \right)$, $P\left( {{X_1}|{S_1}} \right)$ and $P\left( {{X_2}|{S_1}} \right)$ are defined on the Bayesian network in Fig.\ref{fig_2}. $S_{_1}^i,S_{_2}^i\sim P\left( {{S_1},{S_2}} \right)$, then for arbitrary small number ${{\varepsilon}}$,\\
\[\begin{array}{l}
\mathop {\lim }\limits_{N \to \infty } P\left( {\left| {\sum\limits_{i = 1}^N {P\left( {C = 1|S_{_1}^i,S_{_2}^i} \right)\frac{{P\left( {{x_1},{x_2}|S_{_1}^i,S_{_2}^i} \right)}}{{\sum\limits_{i = 1}^N {P\left( {{x_1},{x_2}|S_{_1}^i,S_{_2}^i} \right)} }} - P\left( {C = 1|{x_1},{x_2}} \right)} } \right| < \varepsilon } \right) = 1
\end{array}\]

\textbf{Proof:} Supposing that

${f_1}\left( {{x_1},{x_2}} \right) = \frac{{\sum\limits_{i = 1}^N {P\left( {C = 1|S_{_1}^i,S_{_2}^i} \right)P\left( {{x_1},{x_2}|S_{_1}^i,S_{_2}^i} \right)} }}{{\sum\limits_{i = 1}^N {P\left( {{x_1},{x_2}|S_{_1}^i,S_{_2}^i} \right)} }}$, ${f_2}\left( {{x_1},{x_2}} \right) = \frac{{\sum\limits_{i = 1}^N {P\left( {C = 1|S_{_1}^i,S_{_2}^i} \right)P\left( {{x_1},{x_2}|S_{_1}^i,S_{_2}^i} \right)} }}{{\sum\limits_{j = 1}^N {P\left( {{x_1},{x_2}|\tilde S_{_1}^j,\tilde S_{_2}^j} \right)} }}$, where $S_{_1}^i,S_{_2}^i \sim P\left( {{S_1},{S_2}} \right)$ and $\tilde S_{_1}^j,\tilde S_{_2}^j \sim P\left( {{S_1},{S_2}} \right)$, then

\[\begin{array}{l}
\;\;E\left( {\frac{1}{N}\sum\limits_{i = 1}^N {P\left( {{x_1},{x_2}|S_{_1}^i,S_{_2}^i} \right)} } \right)\\
 = \frac{1}{N}\sum\limits_{i = 1}^N {E\left( {P\left( {{x_1},{x_2}|S_{_1}^i,S_{_2}^i} \right)} \right)} \\
 = \frac{1}{N}\sum\limits_{i = 1}^N {\int\limits_{S_{_1}^i,S_{_2}^i} {P\left( {{x_1},{x_2}|S_{_1}^i,S_{_2}^i} \right)P\left( {S_{_1}^i,S_{_2}^i} \right)} \;dS_{_1}^i,S_{_2}^i} \\
 = P\left( {{x_1},{x_2}} \right),
\end{array}\]

\[\begin{array}{l}
\;\;E\left( {\frac{1}{N}\sum\limits_{j = 1}^N {P\left( {{x_1},{x_2}|\tilde S_{_1}^j,\tilde S_{_2}^j} \right)} } \right) = P\left( {{x_1},{x_2}} \right),
\end{array}\]

\[\begin{array}{l}
\;\;\;Var\left( {\frac{1}{N}\sum\limits_{i = 1}^N {P\left( {{x_1},{x_2}|S_{_1}^i,S_{_2}^i} \right)} } \right)\;\\
 = \frac{1}{{{N^2}}}\sum\limits_{i = 1}^N {Var\left( {P\left( {{x_1},{x_2}|S_{_1}^i,S_{_2}^i} \right)} \right)} \\
 = \frac{1}{{{N^2}}}\sum\limits_{i = 1}^N {\left( {E\left( {P{{\left( {{x_1},{x_2}|S_{_1}^i,S_{_2}^i} \right)}^2}} \right) - E{{\left( {P\left( {{x_1},{x_2}|S_{_1}^i,S_{_2}^i} \right)} \right)}^2}} \right)} \\
 = \frac{1}{N}\int\limits_{{S_1},{S_2}} {P{{\left( {{x_1},{x_2}|{S_1},{S_2}} \right)}^2}P\left( {{S_1},{S_2}} \right)d{S_1},{S_2} - P{{\left( {{x_1},{x_2}} \right)}^2}} \\
 = \sigma _1^2,
\end{array}\]

\[\begin{array}{l}
\;\;\;Var\left( {\frac{1}{N}\sum\limits_{j = 1}^N {P\left( {{x_1},{x_2}|\tilde S_{_1}^j,\tilde S_{_2}^j} \right)} } \right)\; = \sigma _1^2.
\end{array}\]
Note that the variance is denoted as $\sigma _1^2$. It is easy to use Lemma 2 to show that for arbitrary small number $\varepsilon $,

\[\begin{array}{l}
P\left( {\left| {\frac{1}{{\frac{1}{N}\sum\limits_{j = 1}^N {P\left( {{x_1},{x_2}|\tilde S_{_1}^j,\tilde S_{_2}^j} \right)} }} - \frac{1}{{P\left( {{x_1},{x_2}} \right)}}} \right| < \varepsilon } \right) \ge 1 - \frac{{\sigma _1^2}}{{NP{{\left( {{x_1},{x_2}} \right)}^2}{\varepsilon ^2}}},
\end{array}\]

\[\begin{array}{l}
\;P\left( {\left| {\frac{1}{{\frac{1}{N}\sum\limits_{i = 1}^N {P\left( {{x_1},{x_2}|S_{_1}^i,S_{_2}^i} \right)} }} - \frac{1}{{P\left( {{x_1},{x_2}} \right)}}} \right| < \varepsilon } \right) \ge 1 - \frac{{\sigma _1^2}}{{NP{{\left( {{x_1},{x_2}} \right)}^2}{\varepsilon ^2}}}.
\end{array}\]
The equations above indicate that

\[\begin{array}{l}
P\left( {\left| {\frac{1}{{\frac{1}{N}\sum\limits_{j = 1}^N {P\left( {{x_1},{x_2}|\tilde S_{_1}^j,\tilde S_{_2}^j} \right)} }} - \frac{1}{{\frac{1}{N}\sum\limits_{i = 1}^N {P\left( {{x_1},{x_2}|S_{_1}^i,S_{_2}^i} \right)} }}} \right| < 2\varepsilon } \right) \ge {\left( {1 - \frac{{\sigma _1^2}}{{NP{{\left( {{x_1},{x_2}} \right)}^2}{\varepsilon ^2}}}} \right)^2}.
\end{array}\]
As $\left| {{f_1}\left( {{x_1},{x_2}} \right) - {f_2}\left( {{x_1},{x_2}} \right)} \right| \le \left| {\frac{1}{{\frac{1}{N}\sum\limits_{j = 1}^N {P\left( {{x_1},{x_2}|\tilde S_{_1}^j,\tilde S_{_2}^j} \right)} }} - \frac{1}{{\frac{1}{N}\sum\limits_{i = 1}^N {P\left( {{x_1},{x_2}|S_{_1}^i,S_{_2}^i} \right)} }}} \right|$, we can get for arbitrary small number $\varepsilon $,

\[\begin{array}{l}
P\left( {\left| {{f_1}\left( {{x_1},{x_2}} \right) - {f_2}\left( {{x_1},{x_2}} \right)} \right| < 2\varepsilon } \right) \ge {\left( {1 - \frac{{\sigma _1^2}}{{NP{{\left( {{x_1},{x_2}} \right)}^2}{\varepsilon ^2}}}} \right)^2}.
\end{array}\]
We also can get that

\[\begin{array}{l}
\;\;E\left( {\frac{1}{N}\sum\limits_{i = 1}^N {P\left( {C = 1|S_{_1}^i,S_{_2}^i} \right)P\left( {{x_1},{x_2}|S_{_1}^i,S_{_2}^i} \right)} } \right)\\
 = \frac{1}{N}\sum\limits_{i = 1}^N {E\left( {P\left( {C = 1|S_{_1}^i,S_{_2}^i} \right)P\left( {{x_1},{x_2}|S_{_1}^i,S_{_2}^i} \right)} \right)} \\
 = \frac{1}{N}\sum\limits_{i = 1}^N {\left\{ {\int\limits_{S_{_1}^i,S_{_2}^i} {\left( {P\left( {C = 1|S_{_1}^i,S_{_2}^i} \right)P\left( {{x_1},{x_2}|S_{_1}^i,S_{_2}^i} \right)} \right.} \;} \right.} \\
\left. {\left. {\;\;\;\;\;\;\;\;\;\;\;\;\;\;\;\;\;\;\;\;\;\;\;\;\;P\left( {S_{_1}^i,S_{_2}^i} \right)dS_{_1}^i,S_{_2}^i} \right)} \right\}\\
 = P\left( {C = 1|{x_1},{x_2}} \right)P\left( {{x_1},{x_2}} \right),
\end{array}\]

\[\begin{array}{l}
\;\;\;Var\left( {\frac{1}{N}\sum\limits_{i = 1}^N {P\left( {C = 1|S_{_1}^i,S_{_2}^i} \right)P\left( {{x_1},{x_2}|S_{_1}^i,S_{_2}^i} \right)} } \right)\;\;\\
 = \frac{1}{N}\left\{ {\int\limits_{{S_1},{S_2}} {\left( {P{{\left( {C = 1|{S_1},{S_2}} \right)}^2}P{{\left( {{x_1},{x_2}|{S_1},{S_2}} \right)}^2}} \right.} } \right.\\
\left. {\;\left. {P\left( {{S_1},{S_2}} \right)d{S_1},{S_2}} \right)\; - P{{\left( {C = 1|{x_1},{x_2}} \right)}^2}P{{\left( {{x_1},{x_2}} \right)}^2}} \right\}\\
 = \sigma _2^2.
\end{array}\]
Note that the variance is denoted as $\sigma _2^2$ here. Since $\frac{{P\left( {C = 1|{x_1},{x_2}} \right)P\left( {{x_1},{x_2}} \right)}}{{P\left( {{x_1},{x_2}} \right)}} = P\left( {C = 1|{x_1},{x_2}} \right)$, it is easy to use Lemma 1 to show that for arbitrary small number $\varepsilon $,

\[\begin{array}{l}
P\left( {\left| {{f_2}\left( {{x_1},{x_2}} \right) - P\left( {C = 1|{x_1},{x_2}} \right)} \right| < \varepsilon } \right) > 1 - \frac{{16\sigma _2^2}}{{NP{{\left( {{x_1},{x_2}} \right)}^2}{\varepsilon ^2}}} - \frac{{16P{{\left( {C = 1|{x_1},{x_2}} \right)}^2}\sigma _1^2}}{{NP{{\left( {{x_1},{x_2}} \right)}^2}{\varepsilon ^2}}},
\end{array}\]
We also know that

\[\begin{array}{l}
P\left( {\left| {{f_1}\left( {{x_1},{x_2}} \right) - {f_2}\left( {{x_1},{x_2}} \right)} \right| < 2\varepsilon } \right) \ge {\left( {1 - \frac{{\sigma _1^2}}{{NP{{\left( {{x_1},{x_2}} \right)}^2}{\varepsilon ^2}}}} \right)^2},
\end{array}\]
then for arbitrary small number $\varepsilon $,

\[\begin{array}{l}
P\left( {\left| {{f_1}\left( {{x_1},{x_2}} \right) - P\left( {C = 1|{x_1},{x_2}} \right)} \right| < 3\varepsilon } \right)\\
 \ge {\left( {1 - \frac{{\sigma _1^2}}{{NP{{\left( {{x_1},{x_2}} \right)}^2}{\varepsilon ^2}}}} \right)^2} - \frac{{16\sigma _2^2}}{{NP{{\left( {{x_1},{x_2}} \right)}^2}{\varepsilon ^2}}} - \frac{{16P{{\left( {C = 1|{x_1},{x_2}} \right)}^2}\sigma _1^2}}{{NP{{\left( {{x_1},{x_2}} \right)}^2}{\varepsilon ^2}}}.
\end{array}\]
This term holds as $P\left( {A \cap B} \right) \ge P\left( A \right) + P\left( B \right) - 1$. Then for arbitrary small number $\varepsilon $, we have

\[\begin{array}{l}
P\left( {\left| {{f_1}\left( {{x_1},{x_2}} \right) - P\left( {C = 1|{x_1},{x_2}} \right)} \right| < \varepsilon } \right)\\
 \ge {\left( {1 - \frac{{9\sigma _1^2}}{{NP{{\left( {{x_1},{x_2}} \right)}^2}{\varepsilon ^2}}}} \right)^2} - \frac{{144\sigma _2^2}}{{NP{{\left( {{x_1},{x_2}} \right)}^2}{\varepsilon ^2}}} - \frac{{144P{{\left( {C = 1|{x_1},{x_2}} \right)}^2}\sigma _1^2}}{{NP{{\left( {{x_1},{x_2}} \right)}^2}{\varepsilon ^2}}}.
\end{array}\]
When $N$ goes to infinite, we calculate the limits for both sides and conclude that, for arbitrary small number $\varepsilon $,

\[\begin{array}{l}
\mathop {\lim }\limits_{N \to \infty } P\left( {\left| {\sum\limits_{i = 1}^N {P\left( {C = 1|S_{_1}^i,S_{_2}^i} \right)\frac{{P\left( {{x_1},{x_2}|S_{_1}^i,S_{_2}^i} \right)}}{{\sum\limits_{i = 1}^N {P\left( {{x_1},{x_2}|S_{_1}^i,S_{_2}^i} \right)} }} - P\left( {C = 1|{x_1},{x_2}} \right)} } \right| < \varepsilon } \right) = 1.
\end{array}\]

\section*{Acknowledgments}
The authors would like to thank Y. Gao, F. Deng and S. Guo for helpful discussion. This work is supported in part by the National Natural Science Foundation of China under Grant 61271388 and Grant 61327902, and in part by the Research Project of Tsinghua University under Grant 2012Z01011 and Grant 20141080934.




\section*{References}
  \bibliographystyle{elsarticle-num}
  \bibliography{neuarvix}





\end{document}